\documentclass[10pt,twocolumn,letterpaper]{article}
\usepackage{indentfirst}
\usepackage{amsfonts}
\usepackage{pifont}
\usepackage{circledsteps}
\usepackage{makecell}
\usepackage{multirow}
\usepackage{booktabs}
\usepackage{graphicx}
\usepackage{caption}
\usepackage{amsmath}
\usepackage{caption}
\usepackage{amsmath}
\usepackage{xcolor}
\usepackage{cancel}

\usepackage[review]{cvpr}      

\usepackage{todonotes}

\definecolor{cvprblue}{rgb}{0.21,0.49,0.74}
\usepackage[pagebackref,breaklinks,colorlinks,citecolor=cvprblue]{hyperref}

\def\paperID{4772} 
\def\confName{CVPR}
\def\confYear{2024}

\author{
  Wentao Qu$^{[1]}$
  \affiliation{Nanjing University of Science and Technology},
  Yuantian Shao$^{[1]}$
 \affiliation{Nanjing University of Science and Technology},
  Lingwu Meng$^{[1]}$
  \affiliation{Nanjing University of Science and Technology},  
  Xiaoshui Huang$^{[2]}$
  \thanks{Corresponding Author 1:}
  \affiliation{Shanghai AI Lab},
  Liang Xiao$^{[1]}$
  \thanks{Corresponding Author 2:}
  \affiliation{Nanjing University of Science and Technology} 
  \\
  $[1]$ Nanjing University of Science and Technology, $[2]$ Shanghai AI Laboratory
}

\nolinenumbers

\date{}
\begin{document}
\title{A Conditional Denoising Diffusion Probabilistic Model for Point Cloud Upsampling}
\maketitle

\begin{abstract}
Point cloud upsampling (PCU) enriches the representation of raw point clouds, significantly improving the performance in downstream tasks such as classification and reconstruction. Most of the existing point cloud upsampling methods focus on sparse point cloud feature extraction and upsampling module design. In a different way, we dive deeper into directly modelling the gradient of data distribution from dense point clouds. In this paper, we proposed a conditional denoising diffusion probability model (DDPM) for point cloud upsampling, called PUDM. Specifically, PUDM treats the sparse point cloud as a condition, and iteratively learns the transformation relationship between the dense point cloud and the noise. Simultaneously, PUDM aligns with a dual mapping paradigm to further improve the discernment of point features. In this context, PUDM enables learning complex geometry details in the ground truth through the dominant features, while avoiding an additional upsampling module design. Furthermore, to generate high-quality arbitrary-scale point clouds during inference, PUDM exploits the prior knowledge of the scale between sparse point clouds and dense point clouds during training by parameterizing a rate factor. Moreover, PUDM exhibits strong noise robustness in experimental results. In the quantitative and qualitative evaluations on PU1K and PUGAN, PUDM significantly outperformed existing methods in terms of Chamfer Distance (CD) and Hausdorff Distance (HD), achieving state of the art (SOTA) performance.
\end{abstract}

\begin{figure}[htp]
	\centering
	\includegraphics[width=0.48\textwidth]{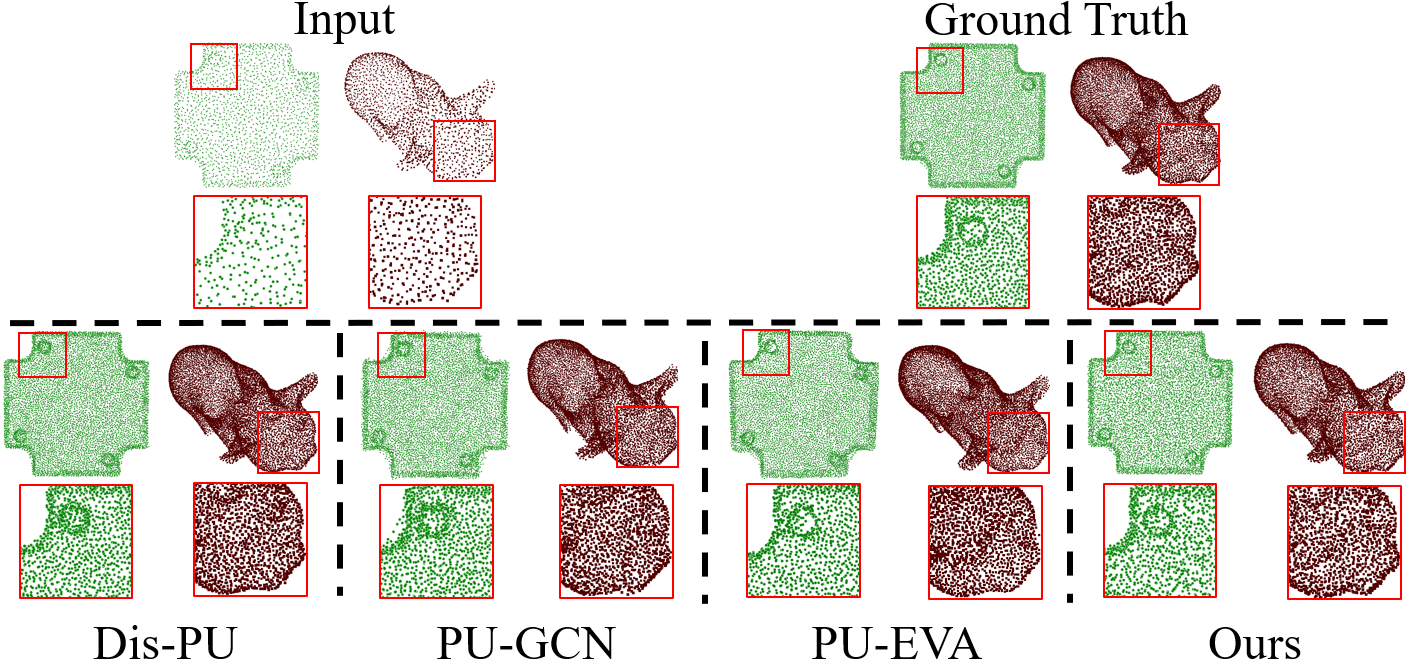}
	\caption{Most existing methods achieving satisfactory results for the input sparse point clouds with clear geometric structures (such as the hole on the green cover rear), but performing poorly for those with fuzzy geometric details (like the eyes of the red pig). Our results, with close proximity to the ground truth.}
	\label{fig_sample}

\end{figure}

\section{Introduction}
Point clouds, as a most fundamental 3D representation, have been widely used in various downstream tasks such as 3D reconstruction \cite{lin2018learning,melas2023pc2}, autonomous driving \cite{cui2021deep,li2021deepi2p,zheng2022global}, and robotics technology \cite{yang2020novel,liu2022weaklabel3d}. However, raw point clouds captured from 
3D sensors
often exhibit sparsity, noise, and non-uniformity. This is substantiated across diverse publicly available benchmark datasets, such as KITTI \cite{geiger2013vision}, ScanNet \cite{dai2017scannet}. Hence, point cloud upsampling, which involves the transformation of sparse, incomplete, and noisy point clouds into denser, complete, and artifact-free representations, has garnered considerable research interest. 


Inspired by deep learning, the pioneering work of PU-Net \cite{yu2018pu} is the first to utilize deep neural networks to address this problem. This first divides the input point cloud into multiple patches and then extracts multi-scale features. Subsequently, these features are aggregated and fed into an upsampling module to approximate the dense point cloud coordinates. Building this approach, for more accurate results, many works \cite{yifan2019patch,li2019pu,li2021point,luo2021pu,qian2021pu} optimize neural networks by focusing on sparse point cloud feature extraction and upsampling module design. 

However, while these methods have achieved improved results, predicting dense point cloud coordinates via sparse point cloud features is an \textbf{indirect} approximating approach. Typically, these methods first utilize an encoder to extract sparse point cloud features, and then use a carefully designed upsampling module to fit dense point cloud coordinates. This approach has three limitations. First, the non-dominance of features causes the generated results to be more inclined toward the input sparse point clouds, struggling to represent reasonable geometry details from the ground truth, as Fig \ref{fig_sample} illustrated. Second, the additional upsampling module designs increase the workload for algorithm designers and often disrupt the intrinsic coordinate mappings in points \cite{yu2018pu,yifan2019patch,qian2021pu}. Third, they mostly require the joint supervision of the CD loss and other losses, making them sensitive to noise \cite{wu2021balanced,huang2022spovt}. 

In this paper, we consider the point cloud upsampling task as a conditional generation problem. This first explores the incorporation of probability models for point cloud upsampling. We propose a novel point cloud upsampling network, called PUDM, which is formally based on a conditional DDPM. Unlike previous methods, PUDM models the gradient of data distribution from dense point clouds (i.e., the ground truth), \textbf{directly} utilizing the dominant features to fit the ground truth, and decoupling the dependency on CD loss. Moreover, the auto-regressive nature of DDPM enables PUDM to efficiently avoid the additional upsampling module design, ensuring intrinsic point-wise mapping relationships in point clouds. 

Simultaneously, to improves the ability of perceiving point features, PUDM employs a dual mapping paradigm. This naturally establishes a dual mapping relationship: between the generated sparse point cloud and the sparse point cloud, and between the dense point cloud and the noise in PUDM. In this context, PUDM has the ability to learn complex geometric structures from the ground truth, generating uniform surfaces aligned with the ground truth, as Fig \ref{fig_sample}.

Furthermore, we found that DDPM only models fixed-scale point cloud objects during training. To overcome the issue, we consider parameterizing a rate factor to exploit the prior knowledge of the scale between sparse point clouds and dense point cloud. In this way, PUDM facilitates the production of high-fidelity arbitrary-scale point clouds during inference.

In additional, benefiting from the inherent denoising architecture and the non-dependency for CD loss, PUDM demonstrates a remarkable degree of robustness in noise experiments. 

Our key contributions can be summarized as: 
\begin{itemize}
    \item We systematically analyze and recognize conditional DDPM as a favorable model for generating uniform point clouds at arbitrary scales in point cloud upsampling tasks.
    \item We propose a novel network with a dual mapping for point cloud upsampling, named PUDM, which is based on conditional DDPM. 
    \item By exploiting the rate prior, PUDM exhibits the ability of generating high-fidelity point clouds across arbitrary scales during inference.
    \item Comprehensive experiments demonstrate the outstanding capability of PUDM in generating geometric details in public benchmarks of point cloud upsampling.
   
\end{itemize}

\section{Related Works}
\textbf{Learnable Point Cloud Upsampling.} The integration of deep learning with formidable data-driven and trainable attributes has markedly accelerated progress within the 3D field. Thanks to the powerful representation capabilities of deep neural networks, directly learning features from 3D data has become achievable, such as PointNet \cite{qi2017pointnet}, PointNet++ \cite{qi2017pointnet++}, DGCNN \cite{phan2018dgcnn}, MinkowskiEngine \cite{choy20194d}, and KPConv \cite{thomas2019kpconv}. PU-Net \cite{yu2018pu} stands as the pioneer in integrating deep neural networks into point cloud upsampling tasks. This first aggregates multi-scale features for each point through multiple MLPs, and then expands them into a point cloud upsampling set via a channel shuffle layer. Following this pattern, some methods have achieved more significant results, such as MPU \cite{yifan2019patch}, PU-GAN \cite{li2019pu}, Dis-PU \cite{li2021point}, and PU-GCN \cite{qian2021pu}. PU-EVA \cite{luo2021pu} is the first to achieve the arbitrary point clouds upsampling rates via edge-vector based affine combinations in one-time training. Subsequently, PUGeo \cite{qian2020pugeo} and NePs \cite{feng2022neural} believe that sampling points within a 2D continuous space can generate higher-quality results. Furthermore, Grad-PU \cite{he2023grad} transforms the point cloud upsampling problem into a coordinate approximation task, avoiding the upsampling module design.

Most methods predict the dense point cloud coordinates via sparse point cloud features, and extend the point set relying on an upsampling module. This causes them to struggle to learn complex geometry details from the ground truth. Moreover, they frequently exhibit a susceptibility to noise due to depending on CD loss during training. In this paper, we consider transforming the point cloud upsampling task into a point cloud generation problem, and first utilize conditional DDPM to address the aforementioned issues. 

\textbf{DDPM for Point Cloud Generation.} Inspired by the success in image generation tasks \cite{ramesh2021zero, ramesh2022hierarchical, rombach2022high}, there has been greater attention on directly generating point clouds through DDPM. \cite{luo2021diffusion} represents the pioneering effort in applying DDPM to unconditional point cloud generation. Subsequently, \cite{zhou20213d} extends the application of DDPM to the point cloud completion task by training a point-voxel CNN \cite{liu2019point}. However, the voxelization process introduces additional computational complexity. Furthermore, PDR \cite{lyu2021conditional} takes raw point clouds as input. But this requires training the two stages (coarse-to-fine) of diffusion models, resulting in a greater time overhead.

In this paper, we explore to the application of conditional DDPM to handle the point cloud upsampling task. Unlike the point cloud generation and completion task, point cloud upsampling exhibits the difference of the point cloud scale between training and inference. We overcome this issue by exploiting a rate prior during training. Meanwhile, our method based on a dual mapping paradigm enables the efficient learning of complex geometric details in a single-stage training.

\section{Denoising Diffusion Probabilistic Model}

\subsection{Background for DDPM}
\label{3.1}

\textbf{The forward and reverse process.} Given the dense point cloud $x$ sampled from a meaningful point distribution $P_{data}$, and an implicit variable $z$ sampled from a tractable noise distribution $P_{latent}$, DDPM establishes the transformation relationship between $x$ and $z$ through two Markov chains. This conducts an auto-regressive process: a forward process $q$ that gradually adds noise to $x$ until $x$ degrades to $z$, and a reverse process $p$ that slowly removes noise from $z$ until $z$ recovers to $x$. We constrain the transformation speed using a time step $t \sim \mathcal{U}(T)$ ($T=1000$ in this paper). 

\textbf{Training objective under specific conditions.} 
Given a set of conditions $C=\{c_i|i=1..S\}$, the training objective of DDPM under specific conditions is (please refer to the supplementary materials for the detailed derivation):

\begin{equation}
\begin{split}
	\label{f311}
	L(\theta) =
 \mathbb{E}_{t \sim U(T), \epsilon \sim \mathcal{N}(0,I)}||\epsilon - \epsilon_\theta(\sqrt{1-\overline{\alpha}_t} \epsilon + \sqrt{\overline{\alpha}_t}x_0,C,t)||^2 
\end{split}
\end{equation}

\textbf{The gradient of data distribution.} Furthermore, we use a stochastic differential equation (SDE) to describe the process of DDPM \cite{song2020score}:

\begin{equation}
\begin{split}
	\label{f312}
	s_\theta(x_t,t)=\nabla_x \log (x_t) =-\frac{1}{\sqrt{1-\overline{\alpha}_t}}\epsilon_\theta(x_t,t)
\end{split}
\end{equation}

The training objective of DDPM is essentially equivalent to computing the score (the gradient of data distribution), which differs only by a constant factor $-\frac{1}{\sqrt{1-\overline{\alpha}_t}}$. 

\subsection{Analysis of DDPM for PCU}
\label{3.2}

We pioneer the exploration of the advantages and limitations of DDPM for PCU, hoping these insights encourage more researchers to introduce probability models into PCU.

\textbf{DDPM is an effective model for PCU.} As mentioned in Sec \ref{3.1}, the auto-regressive nature of DDPM allows it to directly learn geometry details of the ground truth using the dominant features, generating closer-to-truth, fine-grained results. 

Simultaneously, the reverse process of DDPM in PCU is:

\begin{equation}
	\label{f321}
	p_\theta(x_{0:T},c)=p(x_T,c)\prod_{t=1}^{T}p_\theta(x_{t-1}|x_t,c)
\end{equation}
where $c$ means the sparse point cloud sampled from a data distribution $P_c$. According to Eq \ref{f321}, the condition $c$ participates in each step of the reverse process. In fact, this is usually achieved using an additional branch network interacting with the noise network, without intrinsically disrupting the auto-regressive process of DDPM, thus cleverly avoiding to design an additional upsampling module. Moreover, the process naturally defines a one-to-one point-wise mapping relationship between the dense point cloud and the noise, preserving the order of points in the diffusion process.

Furthermore, the efficient denoising architecture and the decoupling of CD loss through the auto-regressive nature significantly support the strong noise robustness of DDPM.

\textbf{The limitations of DDPM in PCU.} While DDPM showcases some advantageous attributes within PCU, it also harbors certain potential limitations:
\begin{itemize}
    \item \textbf{Limitation 1:} The lack of effective prior knowledge for point cloud conditional networks results in their weak feature perception capability \cite{qian2022pix4point,zhang2022pointclip,huang2022frozen}, significantly affecting the final generation results (Tab \ref{t451}). Although some methods \cite{lyu2021conditional} compensate for this problem via a two-stage (coarse-to-fine) training approach, they require a higher training cost. 
    \item \textbf{Limitation 2:} The auto-regressive nature of DDPM provides robust modeling capabilities for fixed-scale objects during training, but it struggles to high-quality sample arbitrary-scale ones during inference (Tab \ref{t452}). Some work treats multi-scale upsampling as multiple tasks \cite{yu2018pu,yifan2019patch,qian2021pu}, but it's not advisable for DDPM due to the excessively high training cost.
\end{itemize}

\section{Methodology}

\subsection{Dual mapping Formulation} 
\label{4.1}
For limitation 1, we adopt a dual mapping paradigm. We first provide a formal exposition of its conception, subsequently delineating the manner in which PUDM aligns with these principles, with a particular emphasis on its role. 

Given two point sets of $X_1=\{x_1^i\in \mathbb{R}^3|i=1..M\}$, and $X_2=\{x_2^i\in \mathbb{R}^3|i=1..N\}$ from different data distributions, a network $f_x$ with a dual-branch ($\{f_1,f_2\} = f_x$) structure, and the corresponding supervision signals for these branches ($\{l_1,l_2\} = l_x$) , if $f_x$ satisfies:

\begin{equation}
	\label{f322}
	Y_1=f_1(X_1), \quad Y_2=f_2(X_2)
\end{equation}
where $Y_1=\{y_1^i\in \mathbb{R}^3|i=1..M\}$, $Y_2=\{y_2^i\in \mathbb{R}^3|i=1..N\}$. $f_x$ can be claimed as a dual mapping network. Eq \ref{f322} means that each element in the original input has one and only one corresponding element in the final output in each branch. Similarly, we can further extend this concept to single mapping network ($\{f_1\}=f_x,\{l_1\}=l_x$), and multiple mapping network ($\{f_i|i=1..n\}=f_x,\{l_i|i=1..n\}=l_x$). 

In PUDM, we only require the conditional network to meet the above condition, because the noise network inherently builds a one-to-one point-wise mapping between input and output \cite{lyu2021conditional}. Specifically, we first force the output $y_c=\{y_c^i\in \mathbb{R}^3|i=1..M\}$ from the conditional network $f_\psi$ to approximate the sparse point cloud $c=\{c^i\in \mathbb{R}^3|i=1..M\}$ coordinates via a MLP, and then optimize the process via the mean squared error loss: 

\begin{equation}
	\label{f322}
	L(\psi)=\mathbb{E}_{c \sim P_c}|| c - f_\psi(c) ||^2
\end{equation}

Formally, this establishes a one-to-one point-wise mapping between the input and output for the conditional network, $y_c = f\psi(c)$.  

For point cloud tasks with unordered structures, this pattern effectively enhances network capability in capturing point features by preserving the ordered relationships between input and output points \cite{choy2019fully,huang2021predator}. Moreover, corresponding supervision signals ensure adequate training for each branch network (Fig \ref{fig_generate_input}), providing an effective strategy to address the challenge of lacking robust 3D pre-trained models as substitutes for conditional branch networks in point cloud generation tasks.

\subsection{Rate Modeling} 
\label{4.2}
For limitation 2, drawing inspiration from the practice of adding class labels in conditional probability models \cite{peebles2023scalable,dhariwal2021diffusion,ho2022classifier}, we propose a simple and effective approach to achieve high-quality arbitrary-scale sampling during inference. Specifically, we first add a rate label $r$ to each sample pair, $(c,x) \rightarrow (c,x,r)$ (the supplementary materials provide ablation studies for different forms of the rate label $r$). Subsequently, we parameter the rate factor using an embedding layer. In this way, the reverse process of DDPM is:

\begin{equation}
	\label{f421}
p_\theta(x_{0:T},c,r)=p(x_T,c,r)\prod_{t=1}^{T}p_\theta(x_{t-1}|x_t,c,r)
\end{equation}

Eq \ref{f421} demonstrates that this simply adds an additional condition to DDPM, the rate prior $r$, without increasing the number of samples. Unlike class labels, we found in experiments that this conditional prior we exploited can significantly improve the generation quality of unseen-scale point clouds. The reason is that generating unseen-scale and seen-category objects usually are easier compared to generating seen-scale and unseen-category ones for models.

\subsection{Network Architecture}
\label{4.3}
In this section, we introduce the overall framework of PUDM, consisting of three crucial components: the conditional network (C-Net), the noise network (N-Net), and the Transfer Module (TM). This process is remarkably illustrated in Fig \ref{fig_pudm}. The parameter setting and implementation details are provided in the supplementary materials.

\begin{figure}[htp]
	\centering
	\includegraphics[width=0.48\textwidth]
 {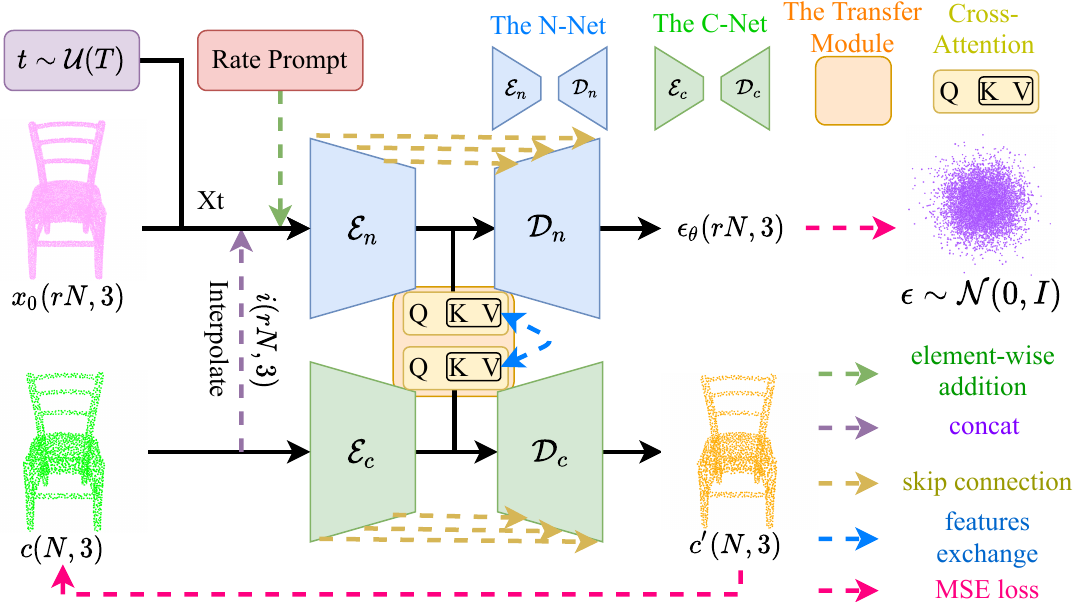}
	\caption{The overall framework of PUDM: The N-Net (upper branch) and the C-Net (lower branch) both establish a one-to-one point-wise mapping between input and output using mean squared error loss. They engage in information exchange through a transfer module (TM). Simultaneously, the rate prompt is provided to exploit the prior knowledge of the scale between sparse point clouds and dense point clouds.}
	\label{fig_pudm}
\end{figure}

\textbf{The Conditional Network (C-Net).} We use PointNet++ \cite{qi2017pointnet++} as the backbone. This follows the standard U-Net framework. The encoder and decoder are composed of multiple Set Abstraction (SA) layers and Feature Propagation (FP) layers, respectively. Unlike PointNet++ using the max-pooling layer to filter features, we consider utilizing the self-attention layer to retain more fine-grained information \cite{zhao2021point,pan2021variational}. In addition, we only feed the sparse point cloud into the C-Net to ensure the feature extraction in a pure and effective manner.

\textbf{The Noise Network (N-Net).} The N-Net and the C-Net share the same network architecture. In contrast to the C-Net, we need to introduce additional guidance information to the N-Net for modeling the diffusion step. 

We first transform the sparse point cloud $c \in \mathbb{R}^{N \times 3}$ into the interpolation point cloud $i \in \mathbb{R}^{rN \times 3}$ through the midpoint interpolation \cite{he2023grad}, and then concatenate $i$ and $x_t$ as the input for the N-Net. Meanwhile, we extract the global features from $i$ to enhance the semantic understanding. Furthermore, to identify the noise level, we encode the time step $t$. Finally, as mentioned in Sec. \ref{4.2}, we parameterized the rate factor $r$. These additional pieces of information are treated as global features, and incorporated into each stage of the encoder and the decoder in the N-Net.

\textbf{The Transfer Module (TM).} We propose a bidirectional interaction module (TM) to serve as an intermediary between the C-Net and the N-Net. We only place the TM at the bottleneck stage of U-Net, due to the significant computational efficiency and the abundant semantic information via the maximum receptive field \cite{huang2021predator,huang2022imfnet}. 

Given the outputs of the encoder in the C-Net and the N-Net, $F^c \in \mathbb{R}^{N_{e}^c \times C_{e}^c}$, $F^n \in \mathbb{R}^{N_{e}^n \times C_{e}^n}$ separately, the TM first transforms $F^c \rightarrow Q \in \mathbb{R}^{N_{e}^c \times C_i}$ and $F^n \rightarrow (K, V) \in \mathbb{R}^{N_{e}^n \times C_i}$ via MLPs. Next, we can obtain the fused feature:

\begin{equation}
    \label{f341}
		F_{f} = MLP(softmax(\frac{Q K^T}{\sqrt{C_{i}}})V)+F^c
\end{equation}

Subsequently, $F_{f}$ is fed into a feed-forward network (FFN) to output the final features. Similarly, the same operation is also applied in reverse direction, so that information flows in both directions, $F^c \rightarrow F^n$ and $F^n \rightarrow F^c$.

\begin{figure*}[htp]
	\centering
	\includegraphics[width=\textwidth]{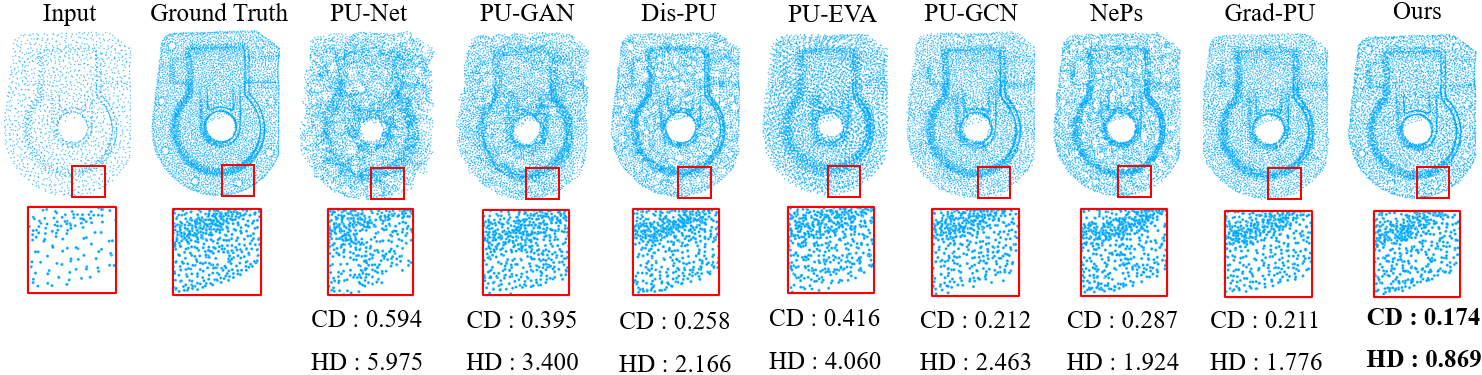}
	\caption{Visualization results at $4 \times$ on PUGAN. Our result exhibits fewer outliers, and clearly captures geometric details from the ground truth (the holes on the casting).}
	\label{fig_outliers}
\end{figure*}

\subsection{Training and Inference}
\label{3.5}
\textbf{Training.} As mentioned earlier (Sec. \ref{4.1} and Sec. \ref{4.2}), PUDM is a dual mapping network, and models the rate prior during training. Therefore, the training objective is:

\begin{equation}
    \label{f351}
		L_{mse} = L(\theta) + \alpha L(\psi)
\end{equation}
where $\alpha$ means a weighting factor ($\alpha = 1$ in this paper). 

\textbf{Inference.} We found that adding the interpolated points $i$ as the guidance information significantly improves the generated quality during inference. Therefore, we iteratively transform $x_t$ into $x_0$ based on :
\begin{equation}
    \label{f352}
x_{t-1}=\gamma(\frac{1}{\sqrt{\alpha_t}}(x_t-\frac{1-a_t}{\sqrt{1-\overline{\alpha}_t}}\epsilon_\theta(x_t,t,c,r))+\sigma_t\epsilon + i)
\end{equation}
where $\gamma$ denotes a scale factor ($\gamma = 0.5$ in this paper). 

\section{Experiments}

\subsection{Experiment Setup}
\textbf{Dataset.} In our experiments, we utilize two public benchmarks (PUGAN \cite{li2019pu}, PU1K \cite{qian2021pu}) for evaluation. We adhere to the official training/testing partitioning protocols for these datasets. This uses Poisson disk sampling \cite{yuksel2015sample} to generate 24,000 and 69,000 uniform patches for training, respectively. Each patch contains 256 points, while the corresponding ground truth has 1024 points. Meanwhile, 27 and 127 point clouds are used for testing, respectively. The input sparse point clouds consist of 2048 points, and are upsampled to $2048 \times R$ points via evaluated methods. 

\textbf{Metrics.} Following \cite{yu2018pu,yifan2019patch,he2023grad}, we employ the Chamfer Distance (CD $\times \; 10^{-3}$), Hausdorff Distance (HD $\times \; 10^{-3}$), and Point-to-Surface Distance (P2F $\times \; 10^{-3}$) as evaluation metrics in our experiments. 

\subsection{Comparison with SOTA}

\textbf{Results on PUGAN.} We first conduct the point cloud upsampling at low upsampling rate ($4\times$) and high upsampling rate ($16\times$) on PUGAN. Table \ref{t421} illustrates the substantial superiority of our method in geometric detail description compared to other methods, as evidenced by significantly reduced CD and HD. Because our method models the gradient of data distribution from dense point clouds, facilitating the direct approximation of geometric details from the ground truth, thereby yielding higher accuracy of our results. Fig \ref{fig_outliers} further substantiates our viewpoint, and shows that our method produces fewer outliers, aligning with more uniform surfaces, finer geometric details, and closer to the ground truth. 

In addition, despite P2F falling slightly behind Grad-PU \cite{he2023grad} at $4\times$,  the difference is insignificant due to the asymmetry between points and surfaces \cite{li2019pu,he2023grad}. 

\begin{table}[h]
        \scriptsize
  \resizebox{0.48\textwidth}{!}{
	\begin{tabular}{p{1.5cm}p{0.6cm}p{0.6cm}p{0.6cm}p{0.005cm}p{0.6cm}p{0.6cm}p{0.6cm}}	
        \bottomrule
  
        \makecell[l]{\multirow{2}{*}{Methods}}
        &\multicolumn{3}{c}{$4\times$} 
        &\quad
        &\multicolumn{3}{c}{$16\times$} \\
         \cline{2-4} \cline{6-8}
       
        &\makecell[c]{CD$\downarrow$}
        &\makecell[c]{HD$\downarrow$}
        &\makecell[c]{P2F$\downarrow$}
        &\quad
        &\makecell[c]{CD$\downarrow$}
        &\makecell[c]{HD$\downarrow$}
        &\makecell[c]{P2F$\downarrow$}\\
       \hline
     
        \makecell[l]{PU-Net \cite{yu2018pu}}
        &\makecell[c]{0.529}
        &\makecell[c]{6.805}
        &\makecell[c]{4.460}
        &\quad
        &\makecell[c]{0.510}
        &\makecell[c]{8.206}
        &\makecell[c]{6.041}\\

        \makecell[l]{MPU \cite{yifan2019patch}}
        &\makecell[c]{0.292}
        &\makecell[c]{6.672}
        &\makecell[c]{2.822}
        &\quad
        &\makecell[c]{0.219}
        &\makecell[c]{7.054}
        &\makecell[c]{3.085}\\
        
        \makecell[l]{PU-GAN \cite{qian2021pu}}
        &\makecell[c]{0.282}
        &\makecell[c]{5.577}
        &\makecell[c]{2.016}
        &\quad
        &\makecell[c]{0.207}
        &\makecell[c]{6.963}
        &\makecell[c]{2.556}\\

        \makecell[l]{Dis-PU \cite{li2021point}}
        &\makecell[c]{0.274}
        &\makecell[c]{3.696}
        &\makecell[c]{1.943}
        &\quad
        &\makecell[c]{0.167}
        &\makecell[c]{4.923}
        &\makecell[c]{2.261}\\

        \makecell[l]{PU-EVA \cite{luo2021pu}}
        &\makecell[c]{0.277}
        &\makecell[c]{3.971}
        &\makecell[c]{2.524}
        &\quad
        &\makecell[c]{0.185}
        &\makecell[c]{5.273}
        &\makecell[c]{2.972}\\

        \makecell[l]{PU-GCN \cite{qian2021pu}}
        &\makecell[c]{0.268}
        &\makecell[c]{3.201}
        &\makecell[c]{2.489}
        &\quad
        &\makecell[c]{0.161}
        &\makecell[c]{4.283}
        &\makecell[c]{2.632}\\

        \makecell[l]{NePS \cite{feng2022neural}}
        &\makecell[c]{0.259}
        &\makecell[c]{3.648}
        &\makecell[c]{1.935}
        &\quad
        &\makecell[c]{0.152}
        &\makecell[c]{4.910}
        &\makecell[c]{2.198}\\

        \makecell[l]{Grad-PU \cite{he2023grad}}
        &\makecell[c]{0.245}
        &\makecell[c]{2.369}
        &\makecell[c]{\textbf{1.893}}
        &\quad
        &\makecell[c]{0.108}
        &\makecell[c]{2.352}
        &\makecell[c]{2.127}\\
        \hline

        \makecell[l]{Ours}
        &\makecell[c]{\textbf{0.131}}
        &\makecell[c]{\textbf{1.220}}
        &\makecell[c]{1.912}
        &\quad
        &\makecell[c]{\textbf{0.082}}
        &\makecell[c]{\textbf{1.120}}
        &\makecell[c]{\textbf{2.114}}\\
        \bottomrule
        
	\end{tabular}
 }
	\caption{The results of $4 \times$ and $16 \times$ on PUGAN. Our method significantly surpasses other methods in terms of CD and HD.}
	\label{t421}
\end{table}

\textbf{Arbitrary Upsampling Rates on PUGAN.} Similarly to \cite{he2023grad}, we perform comparative analyses across various rates on PUGAN. Tab \ref{t422} shows that our method steadily outperforms Grad-PU \cite{he2023grad} across nearly all metrics. In particular, our method demonstrates a significant performance advantage in terms of CD and HD, further affirming the superiority in learning complex geometric details. 

\begin{figure}[htp]
	\centering
	\includegraphics[width=0.45\textwidth]{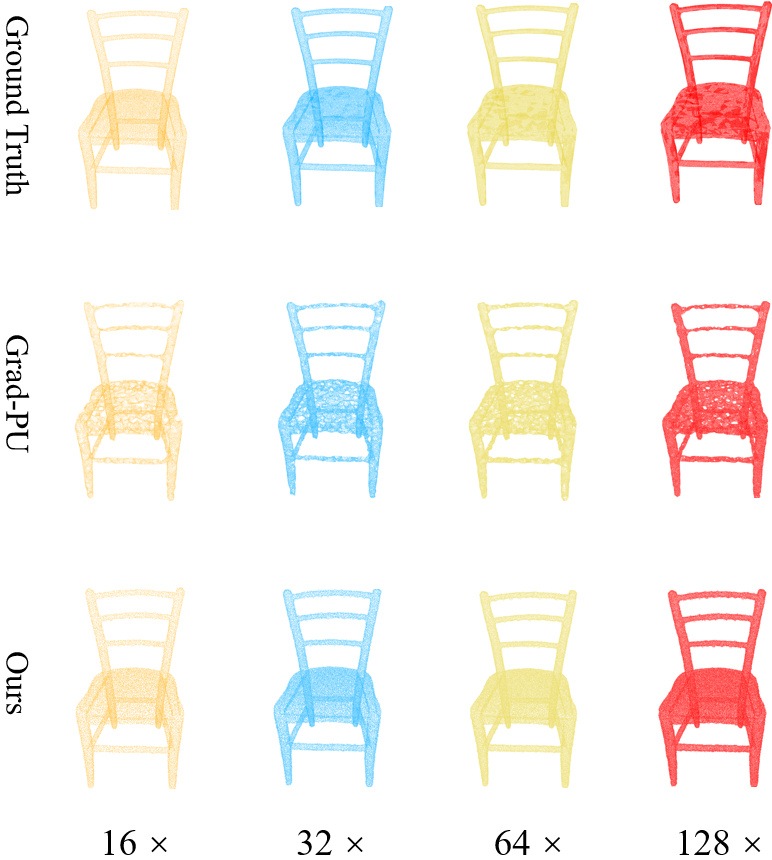}
	\caption{Grad-PU $vs.$ ours at large rates on PUGAN. Our method consistently generates more uniform and smooth surfaces (these results are achieved using an NVIDIA 3090 GPU).}
	\label{fig_big_upsampling}
\end{figure}

Moreover, we visualize the results at higher upsampling rates ($16\times$, $32\times$, $64\times$, and $128\times$) in Fig \ref{fig_big_upsampling}. Our results obviously exhibit more complete, uniform, and smooth compared to Grad-PU \cite{he2023grad}.

\textbf{Results on PU1K.} Furthermore, we also conduct the evaluation at $4\times$ on more challenging PU1K \cite{qian2021pu}. As reported in Tab \ref{t423}, our method continues to demonstrate substantial advantages in terms of CD and HD compared to other methods.

\begin{table}[h]
        \scriptsize
  \resizebox{0.46\textwidth}{!}{
	\begin{tabular}{p{0.8cm}p{0.6cm}p{0.6cm}p{0.6cm}p{0.005cm}p{0.6cm}p{0.6cm}p{0.6cm}}	
        \bottomrule
  
        \makecell[c]{\multirow{2}{*}{Rates}}
        &\multicolumn{3}{c}{Grad-PU \cite{he2023grad}} 
        &\quad
        &\multicolumn{3}{c}{Ours} \\
         \cline{2-4} \cline{6-8}

        &\makecell[c]{CD$\downarrow$}
        &\makecell[c]{HD$\downarrow$}
        &\makecell[c]{P2F$\downarrow$}
        &\quad
        &\makecell[c]{CD$\downarrow$}
        &\makecell[c]{HD$\downarrow$}
        &\makecell[c]{P2F$\downarrow$}\\
       \hline
     
        \makecell[c]{$2 \times$}
        &\makecell[c]{0.540}
        &\makecell[c]{3.177}
        &\makecell[c]{\textbf{1.775}}
        &\quad
        &\makecell[c]{\textbf{0.247}}
        &\makecell[c]{\textbf{1.410}}
        &\makecell[c]{1.812}\\

        \makecell[c]{$3 \times$}
        &\makecell[c]{0.353}
        &\makecell[c]{2.608}
        &\makecell[c]{\textbf{1.654}}
        &\quad
        &\makecell[c]{\textbf{0.171}}
        &\makecell[c]{\textbf{1.292}}
        &\makecell[c]{1.785}\\

        \makecell[c]{$5 \times$}
        &\makecell[c]{0.234}
        &\makecell[c]{2.549}
        &\makecell[c]{1.836}
        &\quad
        &\makecell[c]{\textbf{0.116}}
        &\makecell[c]{\textbf{1.244}}
        &\makecell[c]{\textbf{1.794}}\\

        \makecell[c]{$6 \times$}
        &\makecell[c]{0.225}
        &\makecell[c]{2.526}
        &\makecell[c]{1.981}
        &\quad
        &\makecell[c]{\textbf{0.107}}
        &\makecell[c]{\textbf{1.235}}
        &\makecell[c]{\textbf{1.980}}\\

        \makecell[c]{$7 \times$}
        &\makecell[c]{0.219}
        &\makecell[c]{2.634}
        &\makecell[c]{\textbf{1.940}}
        &\quad
        &\makecell[c]{\textbf{0.106}}
        &\makecell[c]{\textbf{1.231}}
        &\makecell[c]{1.952}\\

        \bottomrule
        
	\end{tabular}
 }
	\caption{Grad-PU $vs.$ ours at various rates on PUGAN. Benefiting from the rate modeling, our method still exhibits remarkable performance at various rates.}
	\label{t422}

\end{table}

\begin{table}[h]
        \scriptsize
  \resizebox{0.48\textwidth}{!}{
	\begin{tabular}{p{1.8cm}p{1.6cm}p{1.6cm}p{1.6cm}}	
        \bottomrule
  
        \makecell[l]{Methods}
        &\makecell[c]{CD$\downarrow$}
        &\makecell[c]{HD$\downarrow$}
        &\makecell[c]{P2F$\downarrow$}\\
       \hline
     
        \makecell[l]{PU-Net \cite{yu2018pu}}
        &\makecell[c]{1.155}
        &\makecell[c]{15.170}
        &\makecell[c]{4.834}\\

        \makecell[l]{MPU \cite{yifan2019patch}}
        &\makecell[c]{0.935}
        &\makecell[c]{13.327}
        &\makecell[c]{3.511}\\
        
        \makecell[l]{PU-GCN \cite{qian2021pu}}
        &\makecell[c]{0.585}
        &\makecell[c]{7.577}
        &\makecell[c]{2.499}\\

        \makecell[l]{Grad-PU \cite{he2023grad}}
        &\makecell[c]{0.404}
        &\makecell[c]{3.732}
        &\makecell[c]{\textbf{1.474}}\\
        \hline

        \makecell[l]{Ours}
        &\makecell[c]{\textbf{0.217}}
        &\makecell[c]{\textbf{2.164}}
        &\makecell[c]{1.477}\\
        \bottomrule
        
	\end{tabular}
 }
	\caption{The results of $4\times$ on PU1K. We utilize the experimental results from the original paper.Our method outperforms other methods across nearly all metrics.}
	\label{t423}
\end{table}

\begin{figure*}[htp]
	\centering
	\includegraphics[width=\textwidth]{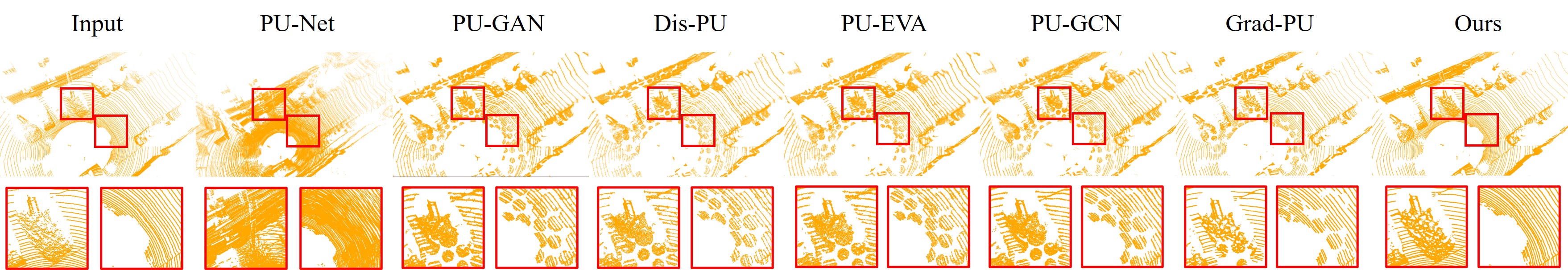}
	\caption{The results of $4\times$ on KITTI. Our method noticeably generates more reasonable and uniform results on some complex geometric structures.}
	\label{fig_KITTI}
\end{figure*}

\begin{figure*}[htp]
	\centering
	\includegraphics[width=\textwidth]{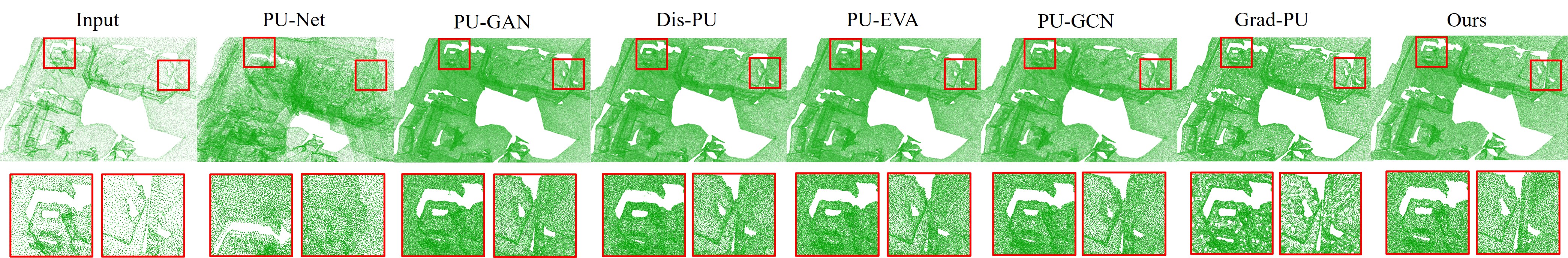}
	\caption{The results of $4\times$ on ScanNet. Our results exhibit reduced instances of outliers, concurrently generating more uniform and complete surfaces.}
	\label{fig_ScanNet}
\end{figure*}

\textbf{Result on Real datasets.}
Additionally, we conduct the evaluation on  real indoor (ScanNet \cite{dai2017scannet}) and outdoor (KITTI \cite{geiger2013vision}) scene datasets. All methods are only trained on PUGAN. Upsampling scene-level point clouds poses greater challenges than upsampling object-level ones, due to the former having more intricate geometric structures. Due to the absence of the ground truth, our analysis is confined to qualitative comparisons. In Fig \ref{fig_KITTI}, our method still generates reasonable and smooth surfaces on some complex structures, while other methods exhibit artifacts such as overlap and voids. Simultaneously, Fig \ref{fig_ScanNet} illustrates that our results show more complete and fewer outliers. Although Grad-PU \cite{he2023grad} also demonstrates good outlier results, it generates a considerable amount of uneven surfaces.

\subsection{Validation for Noise Robustness}
\textbf{Gaussian Noise.} To demonstrate the robustness, we perturb the sparse point clouds with Gaussian noise sampled $\mathcal{N}(0,I)$ added at different noise levels $\tau$. 

As shown in Tab \ref{t424}, our method significantly outperforms other methods under multiple level noise perturbations ($\tau=0.01$, $\tau=0.02$). Specifically, this is because our method models the noise $\epsilon$ (the gradient of data distribution) and avoids CD loss during training.

\begin{table}[h]
        \scriptsize
	\resizebox{0.48\textwidth}{!}{
        \centering
 \begin{tabular}{p{1.5cm}p{0.6cm}p{0.6cm}p{0.6cm}p{0.005cm}p{0.6cm}p{0.7cm}p{0.7cm}}	
        \bottomrule
  
        \makecell[l]{Noise Levels}
        &\multicolumn{3}{c}{$\tau=0.01$} 
        &\quad
        &\multicolumn{3}{c}{$\tau=0.02$} \\
         \cline{2-4} \cline{6-8}

       \makecell[l]{Methods}
        &\makecell[c]{CD$\downarrow$}
        &\makecell[c]{HD$\downarrow$}
        &\makecell[c]{P2F$\downarrow$}
        &\quad
        &\makecell[c]{CD$\downarrow$}
        &\makecell[c]{HD$\downarrow$}
        &\makecell[c]{P2F$\downarrow$}\\
       \hline
     
        \makecell[l]{PU-Net \cite{yu2018pu}}
        &\makecell[c]{0.628}
        &\makecell[c]{8.068}
        &\makecell[c]{9.816}
        &\quad
        &\makecell[c]{1.078}
        &\makecell[c]{10.867}
        &\makecell[c]{16.401}\\

       \makecell[l]{MPU \cite{yifan2019patch}}
        &\makecell[c]{0.506}
        &\makecell[c]{6.978}
        &\makecell[c]{9.059}
        &\quad
        &\makecell[c]{0.929}
        &\makecell[c]{10.820}
        &\makecell[c]{15.621}\\
        
        \makecell[l]{PU-GAN \cite{qian2021pu}}
        &\makecell[c]{0.464}
        &\makecell[c]{6.070}
        &\makecell[c]{7.498}
        &\quad
        &\makecell[c]{0.887}
        &\makecell[c]{10.602}
        &\makecell[c]{15.088}\\

        \makecell[l]{Dis-PU \cite{li2021point}}
        &\makecell[c]{0.419}
        &\makecell[c]{5.413}
        &\makecell[c]{6.723}
        &\quad
        &\makecell[c]{0.818}
        &\makecell[c]{9.345}
        &\makecell[c]{14.376}\\

        \makecell[l]{PU-EVA \cite{luo2021pu}}
        &\makecell[c]{0.459}
        &\makecell[c]{5.377}
        &\makecell[c]{7.189}
        &\quad
        &\makecell[c]{0.839}
        &\makecell[c]{9.325}
        &\makecell[c]{14.652}\\

        \makecell[l]{PU-GCN \cite{qian2021pu}}
        &\makecell[c]{0.448}
        &\makecell[c]{5.586}
        &\makecell[c]{6.989}
        &\quad
        &\makecell[c]{0.816}
        &\makecell[c]{8.604}
        &\makecell[c]{13.798}\\

       \makecell[l]{NePS \cite{feng2022neural}}
        &\makecell[c]{0.425}
        &\makecell[c]{5.438}
        &\makecell[c]{6.546}
        &\quad
        &\makecell[c]{0.798}
        &\makecell[c]{9.102}
        &\makecell[c]{12.088}\\

        \makecell[l]{Grad-PU \cite{he2023grad}}
        &\makecell[c]{0.414}
        &\makecell[c]{4.145}
        &\makecell[c]{6.400}
        &\quad
        &\makecell[c]{0.766}
        &\makecell[c]{7.336}
        &\makecell[c]{11.534}\\
        \hline

        \makecell[l]{Ours}
        &\makecell[c]{\textbf{0.210}}
        &\makecell[c]{\textbf{2.430}}
        &\makecell[c]{\textbf{6.070}}
        &\quad
        &\makecell[c]{\textbf{0.529}}
        &\makecell[c]{\textbf{5.471}}
        &\makecell[c]{\textbf{9.742}}\\
        \bottomrule
        
	\end{tabular}
 }
	\caption{The results of $4\times$ at low-level Gaussian noise on PUGAN. Our method significantly outperforms other methods in terms of noise robustness.}
	\label{t424}
\end{table}

Moreover, we also conduct the evaluation under more challenging noise perturbations. Tab \ref{t425} shows that our method exhibits stronger robustness results at higher level noise perturbations ($\tau=0.05$ and $\tau=0.1$). This indicates that our method exhibits a trend of resilience for the noise robustness.

\begin{table}[h]
        \scriptsize
        \resizebox{0.48\textwidth}{!}{
	\begin{tabular}{p{1.5cm}p{0.6cm}p{0.7cm}p{0.7cm}p{0.005cm}p{0.6cm}p{0.7cm}p{0.7cm}}	
        \bottomrule
  
        \makecell[l]{Noise Levels}
        &\multicolumn{3}{c}{$\tau=0.05$} 
        &\quad
        &\multicolumn{3}{c}{$\tau=0.1$} \\
         \cline{2-4} \cline{6-8}

        \makecell[l]{Methods}
        &\makecell[c]{CD$\downarrow$}
        &\makecell[c]{HD$\downarrow$}
        &\makecell[c]{P2F$\downarrow$}
        &\quad
        &\makecell[c]{CD$\downarrow$}
        &\makecell[c]{HD$\downarrow$}
        &\makecell[c]{P2F$\downarrow$}\\
       \hline
     
        \makecell[l]{PU-Net \cite{yu2018pu}}
        &\makecell[c]{1.370}
        &\makecell[c]{13.729}
        &\makecell[c]{23.249}
        &\quad
        &\makecell[c]{1.498}
        &\makecell[c]{14.193}
        &\makecell[c]{23.846}\\

        \makecell[l]{MPU \cite{yifan2019patch}}
        &\makecell[c]{1.247}
        &\makecell[c]{11.645}
        &\makecell[c]{22.189}
        &\quad
        &\makecell[c]{1.321}
        &\makecell[c]{12.415}
        &\makecell[c]{23.841}\\
        
        \makecell[l]{PU-GAN \cite{qian2021pu}}
        &\makecell[c]{1.124}
        &\makecell[c]{9.091}
        &\makecell[c]{21.252}
        &\quad
        &\makecell[c]{1.271}
        &\makecell[c]{10.911}
        &\makecell[c]{23.174}\\

        \makecell[l]{Dis-PU \cite{li2021point}}
        &\makecell[c]{1.076}
        &\makecell[c]{7.921}
        &\makecell[c]{20.603}
        &\quad
        &\makecell[c]{1.244}
        &\makecell[c]{10.913}
        &\makecell[c]{22.845}\\

        \makecell[l]{PU-EVA \cite{luo2021pu}}
        &\makecell[c]{1.057}
        &\makecell[c]{7.910}
        &\makecell[c]{20.044}
        &\quad
        &\makecell[c]{1.226}
        &\makecell[c]{9.305}
        &\makecell[c]{22.296}\\

        \makecell[l]{PU-GCN \cite{qian2021pu}}
        &\makecell[c]{1.263}
        &\makecell[c]{9.869}
        &\makecell[c]{22.835}
        &\quad
        &\makecell[c]{1.456}
        &\makecell[c]{11.063}
        &\makecell[c]{25.213}\\

        \makecell[l]{NePS \cite{feng2022neural}}
        &\makecell[c]{1.143}
        &\makecell[c]{9.645}
        &\makecell[c]{18.642}
        &\quad
        &\makecell[c]{1.198}
        &\makecell[c]{9.874}
        &\makecell[c]{20.162}\\

        \makecell[l]{Grad-PU \cite{he2023grad}}
        &\makecell[c]{0.978}
        &\makecell[c]{8.057}
        &\makecell[c]{16.927}
        &\quad
        &\makecell[c]{1.118}
        &\makecell[c]{8.946}
        &\makecell[c]{18.845}\\
        \hline

        \makecell[l]{Ours}
        &\makecell[c]{\textbf{0.618}}
        &\makecell[c]{\textbf{5.386}}
        &\makecell[c]{\textbf{14.751}}
        &\quad
        &\makecell[c]{\textbf{0.853}}
        &\makecell[c]{\textbf{6.239}}
        &\makecell[c]{\textbf{16.845}}\\
        \bottomrule
        
	\end{tabular}
 }
	\caption{The results of $4\times$ at high-level Gaussian noise on PUGAN. Compared to other methods, our method demonstrates a more favorable upward trend for robustness to noise.}
	\label{t425}
\end{table}

\textbf{Random Noise.} 
Furthermore, we also investigated the performance of our method on other noise distribution. Admittedly, while our method still keeps SOTA performance, as shown in Tab \ref{t426}, the results on random noise show significantly lower than that on Gaussian noise. 

We provide an intuitive explanation. Eq \ref{f312} demonstrates that the training objective of DDPM is to fit the gradient of data distribution (modeling the noise $\epsilon$, named score) \cite{song2020score}. Essentially, DDPM learns the direction of noise generation. When the conditions with noise are considered, the disturbance in the direction exhibits relatively small, because the noise has a similar distribution to $\epsilon$. Therefore, during inference, our method demonstrates robustness to similar noise distributions of $\epsilon$ (Gaussian noise), but performs poorly when faced with different ones. 

\begin{table}[h]
        \scriptsize
        \resizebox{0.48\textwidth}{!}{
	\begin{tabular}{p{1.5cm}p{0.6cm}p{0.7cm}p{0.7cm}p{0.005cm}p{0.6cm}p{0.7cm}p{0.7cm}}	
        \bottomrule
  
        \makecell[l]{Noise Levels}
        &\multicolumn{3}{c}{$\tau=0.05$} 
        &\quad
        &\multicolumn{3}{c}{$\tau=0.1$} \\
         \cline{2-4} \cline{6-8}

        \makecell[l]{Methods}
        &\makecell[c]{CD$\downarrow$}
        &\makecell[c]{HD$\downarrow$}
        &\makecell[c]{P2F$\downarrow$}
        &\quad
        &\makecell[c]{CD$\downarrow$}
        &\makecell[c]{HD$\downarrow$}
        &\makecell[c]{P2F$\downarrow$}\\
       \hline
     
        \makecell[l]{PU-Net \cite{yu2018pu}}
        &\makecell[c]{1.490}
        &\makecell[c]{14.473}
        &\makecell[c]{23.223}
        &\quad
        &\makecell[c]{1.725}
        &\makecell[c]{15.442}
        &\makecell[c]{25.251}\\

        \makecell[l]{MPU \cite{yifan2019patch}}
        &\makecell[c]{1.224}
        &\makecell[c]{10.842}
        &\makecell[c]{20.456}
        &\quad
        &\makecell[c]{1.545}
        &\makecell[c]{11.645}
        &\makecell[c]{23.512}\\
        
        \makecell[l]{PU-GAN \cite{qian2021pu}}
        &\makecell[c]{1.034}
        &\makecell[c]{7.757}
        &\makecell[c]{18.617}
        &\quad
        &\makecell[c]{1.327}
        &\makecell[c]{9.700}
        &\makecell[c]{21.321}\\

        \makecell[l]{Dis-PU \cite{li2021point}}
        &\makecell[c]{1.006}
        &\makecell[c]{6.856}
        &\makecell[c]{17.873}
        &\quad
        &\makecell[c]{1.314}
        &\makecell[c]{7.463}
        &\makecell[c]{20.980}\\

        \makecell[l]{PU-EVA \cite{luo2021pu}}
        &\makecell[c]{1.024}
        &\makecell[c]{7.534}
        &\makecell[c]{18.179}
        &\quad
        &\makecell[c]{1.334}
        &\makecell[c]{8.056}
        &\makecell[c]{21.158}\\

        \makecell[l]{PU-GCN \cite{qian2021pu}}
        &\makecell[c]{1.045}
        &\makecell[c]{9.643}
        &\makecell[c]{18.899}
        &\quad
        &\makecell[c]{1.325}
        &\makecell[c]{10.877}
        &\makecell[c]{21.633}\\

        \makecell[l]{NePS \cite{feng2022neural}}
        &\makecell[c]{1.048}
        &\makecell[c]{7.345}
        &\makecell[c]{18.054}
        &\quad
        &\makecell[c]{1.321}
        &\makecell[c]{9.645}
        &\makecell[c]{21.314}\\

        \makecell[l]{Grad-PU \cite{he2023grad}}
        &\makecell[c]{1.067}
        &\makecell[c]{6.634}
        &\makecell[c]{17.734}
        &\quad
        &\makecell[c]{1.399}
        &\makecell[c]{7.215}
        &\makecell[c]{21.028}\\
        \hline

        \makecell[l]{Ours}
        &\makecell[c]{\textbf{0.998}}
        &\makecell[c]{\textbf{6.110}}
        &\makecell[c]{\textbf{17.558}}
        &\quad
        &\makecell[c]{\textbf{1.310}}
        &\makecell[c]{\textbf{6.732}}
        &\makecell[c]{\textbf{20.564}}\\
        \bottomrule
        
	\end{tabular}
}
	\caption{The results of $4\times$ at high-level random noise on PUGAN. Our method consistently outperforms other methods on all metrics.}
	\label{t426}
\end{table}

\subsection{Effectiveness in Downstream Task}
We evaluate the effectiveness of upsampling quality in the downstream task: point cloud classification. Meanwhile, we also conducted experiments on point cloud part segmentation, please refer to the supplementary materials.

PointNet \cite{qi2017pointnet} and PointNet++ \cite{qi2017pointnet++} are chosen as the downstream task models due to their significant performance and widespread influence in 3D tasks. We follow the official training and testing procedures. Simultaneously, we select ModelNet40 \cite{wu20153d} (40 categories) and ShapeNet \cite{chang2015shapenet} (16 categories) as the benchmarks for point cloud classification. For a fairer and more effective evaluation, we use only 3D coordinates as the input. Similar to the evaluated strategy on real datasets, all point cloud upsampling methods are only trained on PUGAN.

For evaluation, we first subsample 256/512 points from test point clouds on ModelNet40/ShapeNet. Subsequently, they are upsampled to 1024/2048 points through evaluation methods. As depicted in Tab \ref{t427}, our results significantly improve the classification accuracy compared to the low-res point clouds, and consistently outperforms other methods across all metrics.  

\begin{table}[h]
        \scriptsize
        \resizebox{0.48\textwidth}{!}{
	\begin{tabular}{c|cccc|cccc}	
        \bottomrule

        \makecell[l]{Datasets}
        &\multicolumn{4}{c}{ModelNet40 ($\%$)} \vline
        
        &\multicolumn{4}{c}{ShapeNet ($\%$)} \\
        \cline{2-5} \cline{6-9}
        
        \makecell[l]{Models}
        &\multicolumn{2}{c}{PointNet} 
        &\multicolumn{2}{c}{PointNet++} \vline
        
        &\multicolumn{2}{c}{PointNet} 
        &\multicolumn{2}{c}{PointNet++} \\
        \cline{2-5} \cline{6-9}

        \makecell[l]{Methods}
        &\makecell[c]{IA$\uparrow$}
        &\makecell[c]{CA$\uparrow$}
        &\makecell[c]{IA$\uparrow$}
        &\makecell[c]{CA$\uparrow$}
        
        &\makecell[c]{IA$\uparrow$}
        &\makecell[c]{CA$\uparrow$}
        &\makecell[c]{IA$\uparrow$}
        &\makecell[c]{CA$\uparrow$}\\
       \hline

        \makecell[l]{Low-res}
        &\makecell[c]{87.15}
        &\makecell[c]{83.12}
        &\makecell[c]{88.87}
        &\makecell[c]{84.45}
        
        &\makecell[c]{97.61}
        &\makecell[c]{95.09}
        &\makecell[c]{98.20}
        &\makecell[c]{96.11}\\

        \makecell[l]{High-res}
        &\makecell[c]{90.74}
        &\makecell[c]{87.14}
        &\makecell[c]{92.24}
        &\makecell[c]{89.91}
        
        &\makecell[c]{98.89}
        &\makecell[c]{96.61}
        &\makecell[c]{99.27}
        &\makecell[c]{98.18}\\
        
        \makecell[l]{PU-Net \cite{yu2018pu}}
        &\makecell[c]{88.72}
        &\makecell[c]{85.25}
        &\makecell[c]{88.99}
        &\makecell[c]{85.43}
        
        &\makecell[c]{97.99}
        &\makecell[c]{95.69}
        &\makecell[c]{98.57}
        &\makecell[c]{96.35}\\

        \makecell[l]{MPU \cite{yifan2019patch}}
        &\makecell[c]{89.04}
        &\makecell[c]{85.84}
        &\makecell[c]{89.54}
        &\makecell[c]{86.51}
        
        &\makecell[c]{98.03}
        &\makecell[c]{95.92}
        &\makecell[c]{98.94}
        &\makecell[c]{96.81}\\        
        
        \makecell[l]{PU-GAN \cite{qian2021pu}}
        &\makecell[c]{89.95}
        &\makecell[c]{85.68}
        &\makecell[c]{90.45}
        &\makecell[c]{87.23}

        &\makecell[c]{98.75}
        &\makecell[c]{95.70}
        &\makecell[c]{90.45}
        &\makecell[c]{87.23}\\

        \makecell[l]{Dis-PU \cite{li2021point}}
        &\makecell[c]{88.70}
        &\makecell[c]{85.34}
        &\makecell[c]{89.56}
        &\makecell[c]{86.53}

        &\makecell[c]{98.80}
        &\makecell[c]{96.07}
        &\makecell[c]{99.00}
        &\makecell[c]{97.15}\\

        \makecell[l]{PU-EVA \cite{luo2021pu}}
        &\makecell[c]{89.27}
        &\makecell[c]{85.63}
        &\makecell[c]{89.96}
        &\makecell[c]{86.86}
        
        &\makecell[c]{98.72}
        &\makecell[c]{95.69}
        &\makecell[c]{99.07}
        &\makecell[c]{97.58}\\

        \makecell[l]{PU-GCN \cite{qian2021pu}}
        &\makecell[c]{89.77}
        &\makecell[c]{85.38}
        &\makecell[c]{89.45}
        &\makecell[c]{86.15}

        &\makecell[c]{98.78}
        &\makecell[c]{96.06}
        &\makecell[c]{99.03}
        &\makecell[c]{97.42}\\

        \makecell[l]{NePS \cite{feng2022neural}}
        &\makecell[c]{90.01}
        &\makecell[c]{86.15}
        &\makecell[c]{90.32}
        &\makecell[c]{87.34}

        &\makecell[c]{98.94}
        &\makecell[c]{96.20}
        &\makecell[c]{99.12}
        &\makecell[c]{97.94}\\

        \makecell[l]{Grad-PU \cite{he2023grad}}
        &\makecell[c]{90.05}
        &\makecell[c]{86.06}
        &\makecell[c]{89.98}
        &\makecell[c]{87.49}

        &\makecell[c]{98.82}
        &\makecell[c]{96.19}
        &\makecell[c]{99.10}
        &\makecell[c]{97.63}\\        
        \hline

        \makecell[l]{Ours}
        &\makecell[c]{\textbf{90.33}}
        &\makecell[c]{\textbf{86.54}}
        &\makecell[c]{\textbf{92.14}}
        &\makecell[c]{\textbf{89.42}}

        &\makecell[c]{\textbf{98.85}}
        &\makecell[c]{\textbf{96.58}}
        &\makecell[c]{\textbf{99.13}}
        &\makecell[c]{\textbf{97.99}}\\        
        \bottomrule
        
	\end{tabular}
}
	\caption{The results of point cloud classification. "Low-res" refers to the point cloud subsampled, while "High-res" denotes the original test point cloud. Meanwhile, "IA" stands for instance accuracy, and "CA" denotes class accuracy. Our results have more reasonable, finer-grained, and closer-to-ground truth geometric structures, thereby achieving more significant classification accuracy.}
	\label{t427}
\end{table}

\subsection{Ablation Study} 

\textbf{With/Without the dual mapping paradigm.} Thanks to the rich and structured data, the conditional networks for text or images can  be replaced by powerful pre-trained models \cite{ramesh2022hierarchical,xu2023dream3d,rombach2022high,saharia2022photorealistic}. However, robust pre-trained backbones are lacking in the 3D field due to sparse data and challenging feature extraction \cite{qian2022pix4point,zhang2022pointclip,huang2022frozen}. In this paper, we employ the dual mapping paradigm to augment the capability of perceiving point features within PUDM, ensuring the comprehensive training of the C-Net. To validate this point, we remove the supervision signal from the C-Net to disrupt this pattern. Meanwhile, we also validate the importance of the C-Net by retaining only the N-Net in PUDM .

\begin{figure}[htp]
	\centering
	\includegraphics[width=0.47\textwidth]{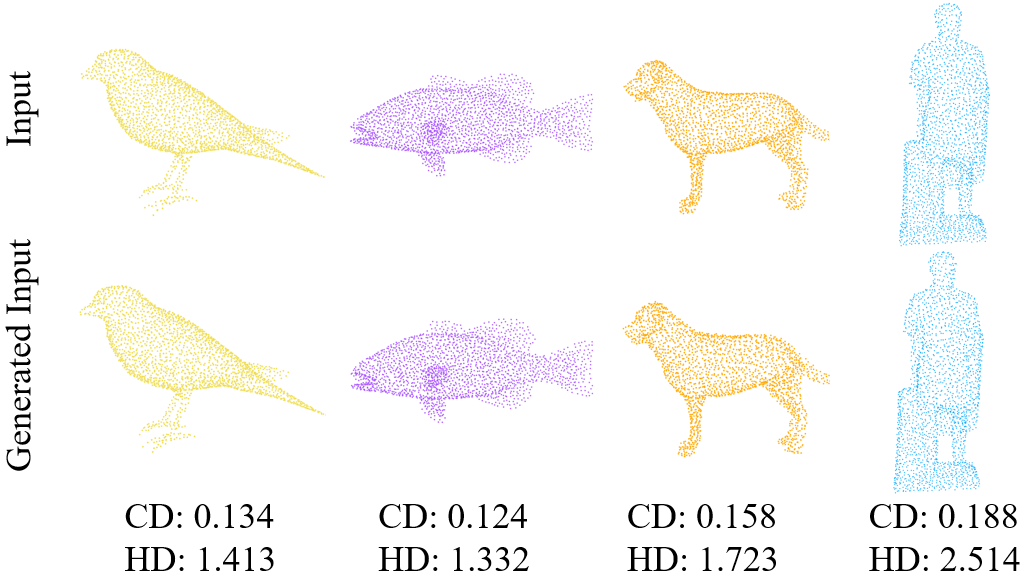}
	\caption{Visualization results of the C-Net generating sparse point clouds on PUGAN. This demonstrates that the C-Net has been effectively trained.}
	\label{fig_generate_input}
\end{figure}

As reported in Tab \ref{t451}, disrupting the dual mapping pattern leads to a significant decrease in performance due to the weakened point feature perception ability of the C-Net. Fig \ref{fig_generate_input} visualizes the results of the C-Net generating input sparse points using the dual mapping paradigm.

Meanwhile, although removing the C-Net can maintain a single mapping pattern, as demonstrated in prior research \cite{yu2018pu,qian2021pu,luo2021pu}, sparse point cloud feature extraction plays a pivotal role in PCU. 

\begin{table}[h]
        \scriptsize
        \resizebox{0.48\textwidth}{!}{
	\begin{tabular}{p{3.0cm}p{1.1cm}p{1.1cm}p{1.1cm}}	
        \bottomrule
  
        \makecell[l]{Methods}
        &\makecell[c]{CD$\downarrow$}
        &\makecell[c]{HD$\downarrow$}
        &\makecell[c]{P2F$\downarrow$}\\
       \hline

        \makecell[l]{Without the C-Net}
        &\makecell[c]{0.212}
        &\makecell[c]{2.015}
        &\makecell[c]{2.284}\\
        
        \makecell[l]{Without the dual mapping}
        &\makecell[c]{0.168}
        &\makecell[c]{1.498}
        &\makecell[c]{2.013}\\

        \makecell[l]{With the dual mapping}
        &\makecell[c]{\textbf{0.131}}
        &\makecell[c]{\textbf{1.220}}
        &\makecell[c]{\textbf{1.912}}\\
        \bottomrule
        
	\end{tabular}
 }
	\caption{Ablation study of the dual mapping paradigm. The construction of the dual mapping pattern evidently achieves the best performance.}
	\label{t451}
\end{table}

\textbf{With/Without the rate prior.} As mentioned in Sec \ref{4.2}, we introduce the rate prior into PUDM during training to achieve high-quality generation of point clouds during inference. Tab \ref{t452} demonstrates the effectiveness of this approach. Without the rate prior, the overall performance notably decreases, and exhibits significant fluctuations (performing better at $4 \times$, but worse at other rates). 

\begin{table}[h]
        \scriptsize
 \resizebox{0.48\textwidth}{!}{
	\begin{tabular}{p{0.8cm}p{0.6cm}p{0.7cm}p{0.7cm}p{0.005cm}p{0.6cm}p{0.7cm}p{0.7cm}}	
        \bottomrule
  
        \makecell[c]{\multirow{2}{*}{Rates}}
        &\multicolumn{3}{c}{Without the rate modeling} 
        &\quad
        &\multicolumn{3}{c}{With the rate modeling} \\
         \cline{2-4} \cline{6-8}

        &\makecell[c]{CD$\downarrow$}
        &\makecell[c]{HD$\downarrow$}
        &\makecell[c]{P2F$\downarrow$}
        &\quad
        &\makecell[c]{CD$\downarrow$}
        &\makecell[c]{HD$\downarrow$}
        &\makecell[c]{P2F$\downarrow$}\\
       \hline
     
        \makecell[c]{$2 \times$}
        &\makecell[c]{0.295}
        &\makecell[c]{1.816}
        &\makecell[c]{2.014}
        &\quad
        &\makecell[c]{\textbf{0.247}}
        &\makecell[c]{\textbf{1.410}}
        &\makecell[c]{\textbf{1.812}}\\

        \makecell[c]{$3 \times$}
        &\makecell[c]{0.224}
        &\makecell[c]{1.544}
        &\makecell[c]{1.975}
        &\quad
        &\makecell[c]{\textbf{0.171}}
        &\makecell[c]{\textbf{1.292}}
        &\makecell[c]{\textbf{1.785}}\\

        \makecell[c]{$4 \times$}
        &\makecell[c]{0.158}
        &\makecell[c]{1.512}
        &\makecell[c]{\textbf{1.815}}
        &\quad
        &\makecell[c]{\textbf{0.131}}
        &\makecell[c]{\textbf{1.220}}
        &\makecell[c]{1.912}\\

        \makecell[c]{$5 \times$}
        &\makecell[c]{0.166}
        &\makecell[c]{1.548}
        &\makecell[c]{1.944}
        &\quad
        &\makecell[c]{\textbf{0.116}}
        &\makecell[c]{\textbf{1.244}}
        &\makecell[c]{\textbf{1.794}}\\

        \makecell[c]{$6 \times$}
        &\makecell[c]{0.151}
        &\makecell[c]{1.528}
        &\makecell[c]{\textbf{1.956}}
        &\quad
        &\makecell[c]{\textbf{0.107}}
        &\makecell[c]{\textbf{1.235}}
        &\makecell[c]{1.980}\\

        \makecell[c]{$7 \times$}
        &\makecell[c]{0.144}
        &\makecell[c]{1.425}
        &\makecell[c]{1.988}
        &\quad
        &\makecell[c]{\textbf{0.106}}
        &\makecell[c]{\textbf{1.231}}
        &\makecell[c]{\textbf{1.952}}\\

        \makecell[c]{$8 \times$}
        &\makecell[c]{0.139}
        &\makecell[c]{1.399}
        &\makecell[c]{1.921}
        &\quad
        &\makecell[c]{\textbf{0.104}}
        &\makecell[c]{\textbf{1.215}}
        &\makecell[c]{\textbf{1.875}}\\

        \bottomrule
        
	\end{tabular}
 }
	\caption{Ablation study of the rate prior. Utilizing the rate prior significantly enhances the quality of arbitrary-scale sampling.}
	\label{t452}

\end{table}

\textbf{Single/Multiple Transfer Module.} In this paper, we employ a TM positioned at the bottleneck stage of the U-Net, as its maximum receptive field provides ample contextual information \cite{huang2021predator,huang2022imfnet}. Meanwhile, we also attempt to place multiple TMs at each stage in U-Net to enable the interaction of multi-scale information \cite{lyu2021conditional}. Tab \ref{t454} shows that although multiple TMs lead to a slight improvement in terms of CD loss, it is not cost-effective due to the significant increase in computational cost.

\begin{table}[h]
        \scriptsize
  \resizebox{0.48\textwidth}{!}{
	\begin{tabular}{p{1.8cm}p{1.1cm}p{1.1cm}p{1.1cm}p{1.2cm}}	
        \bottomrule
  
        \makecell[l]{Methods}
        &\makecell[c]{CD$\downarrow$}
        &\makecell[c]{HD$\downarrow$}
        &\makecell[c]{P2F$\downarrow$}
        &\makecell[c]{Params$\downarrow$}\\
       \hline

        \makecell[l]{Multiple TMs}
        &\makecell[c]{\textbf{0.129}}
        &\makecell[c]{1.235}
        &\makecell[c]{1.953}
        &\makecell[c]{28.65M}\\
        
        \makecell[l]{Single TM}
        &\makecell[c]{0.131}
        &\makecell[c]{\textbf{1.220}}
        &\makecell[c]{\textbf{1.912}}
        &\makecell[c]{\textbf{16.03M}}\\

        \bottomrule
        
	\end{tabular}
 }
	\caption{Ablation study of the Transfer Module. Using the single TM clearly strikes a balance between performance and computational cost.}
	\label{t454}
\end{table}

\section{Conclution}
In this paper, we systematically analyze and identify the potential of DDPM as a promising model for PCU. Meanwhile, we propose PUDM based on conditional DDPM. PUDM enables to directly utilize the dominant features to generate geometric details approximating the ground truth. Additionally, we analyze the limitations of applying DDPM to PCU (the absence of efficient prior knowledge for the conditional network and the fixed-scale object modeling), and propose corresponding solutions (a dual mapping paradigm and the rate modeling). Moreover, we offer a straightforward explanation regarding the robustness to noise for PUDM observed in experiments.

{
    \small
    \bibliographystyle{ieeenat_fullname}
    \bibliography{main}

\begin{thebibliography}{51}
\providecommand{\natexlab}[1]{#1}
\providecommand{\url}[1]{\texttt{#1}}
\expandafter\ifx\csname urlstyle\endcsname\relax
  \providecommand{\doi}[1]{doi: #1}\else
  \providecommand{\doi}{doi: \begingroup \urlstyle{rm}\Url}\fi

\bibitem[Chang et~al.(2015)Chang, Funkhouser, Guibas, Hanrahan, Huang, Li, Savarese, Savva, Song, Su, et~al.]{chang2015shapenet}
Angel~X Chang, Thomas Funkhouser, Leonidas Guibas, Pat Hanrahan, Qixing Huang, Zimo Li, Silvio Savarese, Manolis Savva, Shuran Song, Hao Su, et~al.
\newblock Shapenet: An information-rich 3d model repository.
\newblock \emph{arXiv preprint arXiv:1512.03012}, 2015.

\bibitem[Choy et~al.(2019{\natexlab{a}})Choy, Gwak, and Savarese]{choy20194d}
Christopher Choy, JunYoung Gwak, and Silvio Savarese.
\newblock 4d spatio-temporal convnets: Minkowski convolutional neural networks.
\newblock In \emph{Proceedings of the IEEE/CVF conference on computer vision and pattern recognition}, pages 3075--3084, 2019{\natexlab{a}}.

\bibitem[Choy et~al.(2019{\natexlab{b}})Choy, Park, and Koltun]{choy2019fully}
Christopher Choy, Jaesik Park, and Vladlen Koltun.
\newblock Fully convolutional geometric features.
\newblock In \emph{Proceedings of the IEEE/CVF international conference on computer vision}, pages 8958--8966, 2019{\natexlab{b}}.

\bibitem[Cui et~al.(2021)Cui, Chen, Chu, Chen, Tian, Li, and Cao]{cui2021deep}
Yaodong Cui, Ren Chen, Wenbo Chu, Long Chen, Daxin Tian, Ying Li, and Dongpu Cao.
\newblock Deep learning for image and point cloud fusion in autonomous driving: A review.
\newblock pages 722--739. IEEE, 2021.

\bibitem[Dai et~al.(2017)Dai, Chang, Savva, Halber, Funkhouser, and Nie{\ss}ner]{dai2017scannet}
Angela Dai, Angel~X Chang, Manolis Savva, Maciej Halber, Thomas Funkhouser, and Matthias Nie{\ss}ner.
\newblock Scannet: Richly-annotated 3d reconstructions of indoor scenes.
\newblock In \emph{Proceedings of the IEEE conference on computer vision and pattern recognition}, pages 5828--5839, 2017.

\bibitem[Dhariwal and Nichol(2021)]{dhariwal2021diffusion}
Prafulla Dhariwal and Alexander Nichol.
\newblock Diffusion models beat gans on image synthesis.
\newblock \emph{Advances in neural information processing systems}, 34:\penalty0 8780--8794, 2021.

\bibitem[Feng et~al.(2022)Feng, Li, Cai, Luo, and Zhang]{feng2022neural}
Wanquan Feng, Jin Li, Hongrui Cai, Xiaonan Luo, and Juyong Zhang.
\newblock Neural points: Point cloud representation with neural fields for arbitrary upsampling.
\newblock In \emph{Proceedings of the IEEE/CVF Conference on Computer Vision and Pattern Recognition}, pages 18633--18642, 2022.

\bibitem[Geiger et~al.(2013)Geiger, Lenz, Stiller, and Urtasun]{geiger2013vision}
Andreas Geiger, Philip Lenz, Christoph Stiller, and Raquel Urtasun.
\newblock Vision meets robotics: The kitti dataset.
\newblock pages 1231--1237. Sage Publications Sage UK: London, England, 2013.

\bibitem[He et~al.(2023)He, Tang, Zhang, Xue, and Fu]{he2023grad}
Yun He, Danhang Tang, Yinda Zhang, Xiangyang Xue, and Yanwei Fu.
\newblock Grad-pu: Arbitrary-scale point cloud upsampling via gradient descent with learned distance functions.
\newblock In \emph{Proceedings of the IEEE/CVF Conference on Computer Vision and Pattern Recognition}, pages 5354--5363, 2023.

\bibitem[Ho and Salimans(2022)]{ho2022classifier}
Jonathan Ho and Tim Salimans.
\newblock Classifier-free diffusion guidance.
\newblock \emph{arXiv preprint arXiv:2207.12598}, 2022.

\bibitem[Ho et~al.(2020)Ho, Jain, and Abbeel]{ho2020denoising}
Jonathan Ho, Ajay Jain, and Pieter Abbeel.
\newblock Denoising diffusion probabilistic models.
\newblock \emph{Advances in neural information processing systems}, 33:\penalty0 6840--6851, 2020.

\bibitem[Huang et~al.(2021)Huang, Gojcic, Usvyatsov, Wieser, and Schindler]{huang2021predator}
Shengyu Huang, Zan Gojcic, Mikhail Usvyatsov, Andreas Wieser, and Konrad Schindler.
\newblock Predator: Registration of 3d point clouds with low overlap.
\newblock In \emph{Proceedings of the IEEE/CVF Conference on computer vision and pattern recognition}, pages 4267--4276, 2021.

\bibitem[Huang et~al.(2022{\natexlab{a}})Huang, Hsu, and Wang]{huang2022spovt}
Sheng~Yu Huang, Hao-Yu Hsu, and Frank Wang.
\newblock Spovt: Semantic-prototype variational transformer for dense point cloud semantic completion.
\newblock \emph{Advances in Neural Information Processing Systems}, 35:\penalty0 33934--33946, 2022{\natexlab{a}}.

\bibitem[Huang et~al.(2022{\natexlab{b}})Huang, Li, Qu, He, Zuo, and Ouyang]{huang2022frozen}
Xiaoshui Huang, Sheng Li, Wentao Qu, Tong He, Yifan Zuo, and Wanli Ouyang.
\newblock Frozen clip model is efficient point cloud backbone.
\newblock \emph{arXiv preprint arXiv:2212.04098}, 2022{\natexlab{b}}.

\bibitem[Huang et~al.(2022{\natexlab{c}})Huang, Qu, Zuo, Fang, and Zhao]{huang2022imfnet}
Xiaoshui Huang, Wentao Qu, Yifan Zuo, Yuming Fang, and Xiaowei Zhao.
\newblock Imfnet: Interpretable multimodal fusion for point cloud registration.
\newblock \emph{IEEE Robotics and Automation Letters}, 7\penalty0 (4):\penalty0 12323--12330, 2022{\natexlab{c}}.

\bibitem[Li and Lee(2021)]{li2021deepi2p}
Jiaxin Li and Gim~Hee Lee.
\newblock Deepi2p: Image-to-point cloud registration via deep classification.
\newblock In \emph{Proceedings of the IEEE/CVF Conference on Computer Vision and Pattern Recognition}, pages 15960--15969, 2021.

\bibitem[Li et~al.(2019)Li, Li, Fu, Cohen-Or, and Heng]{li2019pu}
Ruihui Li, Xianzhi Li, Chi-Wing Fu, Daniel Cohen-Or, and Pheng-Ann Heng.
\newblock Pu-gan: a point cloud upsampling adversarial network.
\newblock In \emph{Proceedings of the IEEE/CVF international conference on computer vision}, pages 7203--7212, 2019.

\bibitem[Li et~al.(2021)Li, Li, Heng, and Fu]{li2021point}
Ruihui Li, Xianzhi Li, Pheng-Ann Heng, and Chi-Wing Fu.
\newblock Point cloud upsampling via disentangled refinement.
\newblock In \emph{Proceedings of the IEEE/CVF conference on computer vision and pattern recognition}, pages 344--353, 2021.

\bibitem[Lin et~al.(2018)Lin, Kong, and Lucey]{lin2018learning}
Chen-Hsuan Lin, Chen Kong, and Simon Lucey.
\newblock Learning efficient point cloud generation for dense 3d object reconstruction.
\newblock In \emph{proceedings of the AAAI Conference on Artificial Intelligence}, 2018.

\bibitem[Liu et~al.(2022)Liu, Zhao, Gao, and Chen]{liu2022weaklabel3d}
Kangcheng Liu, Yuzhi Zhao, Zhi Gao, and Ben~M Chen.
\newblock Weaklabel3d-net: A complete framework for real-scene lidar point clouds weakly supervised multi-tasks understanding.
\newblock In \emph{2022 international conference on robotics and automation (ICRA)}, pages 5108--5115. IEEE, 2022.

\bibitem[Liu et~al.(2019)Liu, Tang, Lin, and Han]{liu2019point}
Zhijian Liu, Haotian Tang, Yujun Lin, and Song Han.
\newblock Point-voxel cnn for efficient 3d deep learning.
\newblock \emph{Advances in Neural Information Processing Systems}, 32, 2019.

\bibitem[Luo et~al.(2021)Luo, Tang, Zhou, Wang, and Yang]{luo2021pu}
Luqing Luo, Lulu Tang, Wanyi Zhou, Shizheng Wang, and Zhi-Xin Yang.
\newblock Pu-eva: An edge-vector based approximation solution for flexible-scale point cloud upsampling.
\newblock In \emph{Proceedings of the IEEE/CVF International Conference on Computer Vision}, pages 16208--16217, 2021.

\bibitem[Luo and Hu(2021)]{luo2021diffusion}
Shitong Luo and Wei Hu.
\newblock Diffusion probabilistic models for 3d point cloud generation.
\newblock In \emph{Proceedings of the IEEE/CVF Conference on Computer Vision and Pattern Recognition}, pages 2837--2845, 2021.

\bibitem[Lyu et~al.(2021)Lyu, Kong, Xu, Pan, and Lin]{lyu2021conditional}
Zhaoyang Lyu, Zhifeng Kong, Xudong Xu, Liang Pan, and Dahua Lin.
\newblock A conditional point diffusion-refinement paradigm for 3d point cloud completion.
\newblock \emph{arXiv preprint arXiv:2112.03530}, 2021.

\bibitem[Melas-Kyriazi et~al.(2023)Melas-Kyriazi, Rupprecht, and Vedaldi]{melas2023pc2}
Luke Melas-Kyriazi, Christian Rupprecht, and Andrea Vedaldi.
\newblock Pc2: Projection-conditioned point cloud diffusion for single-image 3d reconstruction.
\newblock In \emph{Proceedings of the IEEE/CVF Conference on Computer Vision and Pattern Recognition}, pages 12923--12932, 2023.

\bibitem[Pan et~al.(2021)Pan, Chen, Cai, Zhang, Zhao, Yi, and Liu]{pan2021variational}
Liang Pan, Xinyi Chen, Zhongang Cai, Junzhe Zhang, Haiyu Zhao, Shuai Yi, and Ziwei Liu.
\newblock Variational relational point completion network.
\newblock In \emph{Proceedings of the IEEE/CVF conference on computer vision and pattern recognition}, pages 8524--8533, 2021.

\bibitem[Peebles and Xie(2023)]{peebles2023scalable}
William Peebles and Saining Xie.
\newblock Scalable diffusion models with transformers.
\newblock In \emph{Proceedings of the IEEE/CVF International Conference on Computer Vision}, pages 4195--4205, 2023.

\bibitem[Phan et~al.(2018)Phan, Le~Nguyen, Nguyen, and Bui]{phan2018dgcnn}
Anh~Viet Phan, Minh Le~Nguyen, Yen Lam~Hoang Nguyen, and Lam~Thu Bui.
\newblock Dgcnn: A convolutional neural network over large-scale labeled graphs.
\newblock \emph{Neural Networks}, 108:\penalty0 533--543, 2018.

\bibitem[Qi et~al.(2017{\natexlab{a}})Qi, Su, Mo, and Guibas]{qi2017pointnet}
Charles~R Qi, Hao Su, Kaichun Mo, and Leonidas~J Guibas.
\newblock Pointnet: Deep learning on point sets for 3d classification and segmentation.
\newblock In \emph{Proceedings of the IEEE conference on computer vision and pattern recognition}, pages 652--660, 2017{\natexlab{a}}.

\bibitem[Qi et~al.(2017{\natexlab{b}})Qi, Yi, Su, and Guibas]{qi2017pointnet++}
Charles~Ruizhongtai Qi, Li Yi, Hao Su, and Leonidas~J Guibas.
\newblock Pointnet++: Deep hierarchical feature learning on point sets in a metric space.
\newblock \emph{Advances in neural information processing systems}, 30, 2017{\natexlab{b}}.

\bibitem[Qian et~al.(2021)Qian, Abualshour, Li, Thabet, and Ghanem]{qian2021pu}
Guocheng Qian, Abdulellah Abualshour, Guohao Li, Ali Thabet, and Bernard Ghanem.
\newblock Pu-gcn: Point cloud upsampling using graph convolutional networks.
\newblock In \emph{Proceedings of the IEEE/CVF Conference on Computer Vision and Pattern Recognition}, pages 11683--11692, 2021.

\bibitem[Qian et~al.(2022)Qian, Zhang, Hamdi, and Ghanem]{qian2022pix4point}
Guocheng Qian, Xingdi Zhang, Abdullah Hamdi, and Bernard Ghanem.
\newblock Pix4point: Image pretrained transformers for 3d point cloud understanding.
\newblock 2022.

\bibitem[Qian et~al.(2020)Qian, Hou, Kwong, and He]{qian2020pugeo}
Yue Qian, Junhui Hou, Sam Kwong, and Ying He.
\newblock Pugeo-net: A geometry-centric network for 3d point cloud upsampling.
\newblock In \emph{European conference on computer vision}, pages 752--769. Springer, 2020.

\bibitem[Ramesh et~al.(2021)Ramesh, Pavlov, Goh, Gray, Voss, Radford, Chen, and Sutskever]{ramesh2021zero}
Aditya Ramesh, Mikhail Pavlov, Gabriel Goh, Scott Gray, Chelsea Voss, Alec Radford, Mark Chen, and Ilya Sutskever.
\newblock Zero-shot text-to-image generation.
\newblock In \emph{International Conference on Machine Learning}, pages 8821--8831. PMLR, 2021.

\bibitem[Ramesh et~al.(2022)Ramesh, Dhariwal, Nichol, Chu, and Chen]{ramesh2022hierarchical}
Aditya Ramesh, Prafulla Dhariwal, Alex Nichol, Casey Chu, and Mark Chen.
\newblock Hierarchical text-conditional image generation with clip latents.
\newblock \emph{arXiv preprint arXiv:2204.06125}, 1\penalty0 (2):\penalty0 3, 2022.

\bibitem[Rombach et~al.(2022)Rombach, Blattmann, Lorenz, Esser, and Ommer]{rombach2022high}
Robin Rombach, Andreas Blattmann, Dominik Lorenz, Patrick Esser, and Bj{\"o}rn Ommer.
\newblock High-resolution image synthesis with latent diffusion models.
\newblock In \emph{Proceedings of the IEEE/CVF conference on computer vision and pattern recognition}, pages 10684--10695, 2022.

\bibitem[Saharia et~al.(2022)Saharia, Chan, Saxena, Li, Whang, Denton, Ghasemipour, Gontijo~Lopes, Karagol~Ayan, Salimans, et~al.]{saharia2022photorealistic}
Chitwan Saharia, William Chan, Saurabh Saxena, Lala Li, Jay Whang, Emily~L Denton, Kamyar Ghasemipour, Raphael Gontijo~Lopes, Burcu Karagol~Ayan, Tim Salimans, et~al.
\newblock Photorealistic text-to-image diffusion models with deep language understanding.
\newblock \emph{Advances in Neural Information Processing Systems}, 35:\penalty0 36479--36494, 2022.

\bibitem[Song et~al.(2020{\natexlab{a}})Song, Meng, and Ermon]{song2020denoising}
Jiaming Song, Chenlin Meng, and Stefano Ermon.
\newblock Denoising diffusion implicit models.
\newblock \emph{arXiv preprint arXiv:2010.02502}, 2020{\natexlab{a}}.

\bibitem[Song et~al.(2020{\natexlab{b}})Song, Sohl-Dickstein, Kingma, Kumar, Ermon, and Poole]{song2020score}
Yang Song, Jascha Sohl-Dickstein, Diederik~P Kingma, Abhishek Kumar, Stefano Ermon, and Ben Poole.
\newblock Score-based generative modeling through stochastic differential equations.
\newblock \emph{arXiv preprint arXiv:2011.13456}, 2020{\natexlab{b}}.

\bibitem[Thomas et~al.(2019)Thomas, Qi, Deschaud, Marcotegui, Goulette, and Guibas]{thomas2019kpconv}
Hugues Thomas, Charles~R Qi, Jean-Emmanuel Deschaud, Beatriz Marcotegui, Fran{\c{c}}ois Goulette, and Leonidas~J Guibas.
\newblock Kpconv: Flexible and deformable convolution for point clouds.
\newblock In \emph{Proceedings of the IEEE/CVF international conference on computer vision}, pages 6411--6420, 2019.

\bibitem[Wu et~al.(2021)Wu, Pan, Zhang, Wang, Liu, and Lin]{wu2021balanced}
Tong Wu, Liang Pan, Junzhe Zhang, Tai Wang, Ziwei Liu, and Dahua Lin.
\newblock Balanced chamfer distance as a comprehensive metric for point cloud completion.
\newblock \emph{Advances in Neural Information Processing Systems}, 34:\penalty0 29088--29100, 2021.

\bibitem[Wu et~al.(2015)Wu, Song, Khosla, Yu, Zhang, Tang, and Xiao]{wu20153d}
Zhirong Wu, Shuran Song, Aditya Khosla, Fisher Yu, Linguang Zhang, Xiaoou Tang, and Jianxiong Xiao.
\newblock 3d shapenets: A deep representation for volumetric shapes.
\newblock In \emph{Proceedings of the IEEE conference on computer vision and pattern recognition}, pages 1912--1920, 2015.

\bibitem[Xu et~al.(2023)Xu, Wang, Cheng, Cao, Shan, Qie, and Gao]{xu2023dream3d}
Jiale Xu, Xintao Wang, Weihao Cheng, Yan-Pei Cao, Ying Shan, Xiaohu Qie, and Shenghua Gao.
\newblock Dream3d: Zero-shot text-to-3d synthesis using 3d shape prior and text-to-image diffusion models.
\newblock In \emph{Proceedings of the IEEE/CVF Conference on Computer Vision and Pattern Recognition}, pages 20908--20918, 2023.

\bibitem[Yang et~al.(2020)Yang, Liu, Peng, and Liang]{yang2020novel}
Lei Yang, Yanhong Liu, Jinzhu Peng, and Zize Liang.
\newblock A novel system for off-line 3d seam extraction and path planning based on point cloud segmentation for arc welding robot.
\newblock \emph{Robotics and Computer-Integrated Manufacturing}, 64:\penalty0 101929, 2020.

\bibitem[Yifan et~al.(2019)Yifan, Wu, Huang, Cohen-Or, and Sorkine-Hornung]{yifan2019patch}
Wang Yifan, Shihao Wu, Hui Huang, Daniel Cohen-Or, and Olga Sorkine-Hornung.
\newblock Patch-based progressive 3d point set upsampling.
\newblock In \emph{Proceedings of the IEEE/CVF Conference on Computer Vision and Pattern Recognition}, pages 5958--5967, 2019.

\bibitem[Yu et~al.(2018)Yu, Li, Fu, Cohen-Or, and Heng]{yu2018pu}
Lequan Yu, Xianzhi Li, Chi-Wing Fu, Daniel Cohen-Or, and Pheng-Ann Heng.
\newblock Pu-net: Point cloud upsampling network.
\newblock In \emph{Proceedings of the IEEE conference on computer vision and pattern recognition}, pages 2790--2799, 2018.

\bibitem[Yuksel(2015)]{yuksel2015sample}
Cem Yuksel.
\newblock Sample elimination for generating poisson disk sample sets.
\newblock In \emph{Computer Graphics Forum}, pages 25--32. Wiley Online Library, 2015.

\bibitem[Zhang et~al.(2022)Zhang, Guo, Zhang, Li, Miao, Cui, Qiao, Gao, and Li]{zhang2022pointclip}
Renrui Zhang, Ziyu Guo, Wei Zhang, Kunchang Li, Xupeng Miao, Bin Cui, Yu Qiao, Peng Gao, and Hongsheng Li.
\newblock Pointclip: Point cloud understanding by clip.
\newblock In \emph{Proceedings of the IEEE/CVF Conference on Computer Vision and Pattern Recognition}, pages 8552--8562, 2022.

\bibitem[Zhao et~al.(2021)Zhao, Jiang, Jia, Torr, and Koltun]{zhao2021point}
Hengshuang Zhao, Li Jiang, Jiaya Jia, Philip~HS Torr, and Vladlen Koltun.
\newblock Point transformer.
\newblock In \emph{Proceedings of the IEEE/CVF international conference on computer vision}, pages 16259--16268, 2021.

\bibitem[Zheng et~al.(2022)Zheng, Li, Yang, and Lu]{zheng2022global}
Yuchao Zheng, Yujie Li, Shuo Yang, and Huimin Lu.
\newblock Global-pbnet: A novel point cloud registration for autonomous driving.
\newblock pages 22312--22319. IEEE, 2022.

\bibitem[Zhou et~al.(2021)Zhou, Du, and Wu]{zhou20213d}
Linqi Zhou, Yilun Du, and Jiajun Wu.
\newblock 3d shape generation and completion through point-voxel diffusion.
\newblock In \emph{Proceedings of the IEEE/CVF International Conference on Computer Vision}, pages 5826--5835, 2021.

\end{thebibliography}
}

\newpage

\setcounter{equation}{0}
\setcounter{section}{0}
\setcounter{figure}{0}
\setcounter{table}{0}

\section{Additional Ablation Study}
We provide additional ablation studies to deepen the analysis and understanding for our method.

\textbf{Rate label form.} Our method achieves high-quality arbitrary-rate sampling during inference by parameterizing a rate factor. Therefore, we explore the impact of different rate label forms for the model performance. We provide additional information regarding the scale of points, such as the number of points. As shown in Tab \ref{t111}, the performance difference for the model is relatively small (the variation $< 0.02$ for CD). This reason is that the rate label is solely modeled to identify the scale difference between the sparse point cloud and the dense point cloud, thus it can not significantly improve the performance of the model.

\begin{table}[h]
        \scriptsize
  \resizebox{0.48\textwidth}{!}{
	\begin{tabular}{p{1.8cm}p{1.1cm}p{1.1cm}p{1.1cm}p{1.2cm}}	
       \bottomrule
  
        \makecell[c]{Rate Label Forms}
        &\makecell[c]{CD$\downarrow$}
        &\makecell[c]{HD$\downarrow$}
        &\makecell[c]{P2F$\downarrow$}\\
       \hline

        \makecell[c]{[256,1024,3]}
        &\makecell[c]{0.143}
        &\makecell[c]{1.258}
        &\makecell[c]{\textbf{1.907}}\\

        \makecell[c]{[256,1024]}
        &\makecell[c]{0.145}
        &\makecell[c]{1.352}
        &\makecell[c]{1.913}\\

        \makecell[c]{[0.256,1.024]}
        &\makecell[c]{0.141}
        &\makecell[c]{1.289}
        &\makecell[c]{1.954}\\
        
        \makecell[c]{[3,]}
        &\makecell[c]{\textbf{0.131}}
        &\makecell[c]{\textbf{1.220}}
        &\makecell[c]{1.912}\\

        \bottomrule
       
	\end{tabular}
}
	\caption{The ablation study of the rate label form at $4 \times$ on PUGAN \cite{li2019pu}. Our method performs optimally, when $r=[3,]$ (we set $r=[0,]$ to represent $1 \times$).}
	\label{t111}
\end{table}

\textbf{Sampling intervals.} Our method is formally based on conditional DDPM, thus inevitably lagging behind existing non-DDPM-based methods in terms of sampling speed. Therefore, we conduct the ablation study between the quality and time of generating point clouds under different sampling intervals. To stay within the framework of DDPM,  we refraine from applying acceleration methods such as DDIM \cite{song2020denoising}. This only changes the sampling interval during inference, without retraining the model. 

Tab \ref{t121} shows the trade-off between the performance and the generating time of our method under different sampling intervals. Surprisingly, although the overall performance of the model decreases with increasing sampling intervals, this does not follow a linear trend, presenting an irregular state. We believe that the performance variation of the model is not only related to the sampling interval but also to the distance from $x_1$ to $x_0$. When the distance of time step between $x_1$ and $x_0$ is smaller, the model performs better, such as the sampling interval $= 12$ (with a distance of 3 time steps between $x_1$ and $x_0$). In other words, when given some close time intervals, we should choose the sampling interval that brings $x_1$ closest to $x_0$, rather than the one that evenly divides the total time steps.

\begin{table}[h]
        \scriptsize
  \resizebox{0.48\textwidth}{!}{
	\begin{tabular}{p{1.2cm}p{1.2cm}p{1.1cm}p{1.1cm}p{1.1cm}p{1.2cm}}	
        \bottomrule
  
        \makecell[c]{Intervals}
        &\makecell[c]{Distance}
        &\makecell[c]{CD$\downarrow$}
        &\makecell[c]{HD$\downarrow$}
        &\makecell[c]{P2F$\downarrow$}
        &\makecell[c]{Times(s)$\downarrow$}\\
       \hline

        \makecell[c]{50}
        &\makecell[c]{49}
        &\makecell[c]{0.320}
        &\makecell[c]{2.738}
        &\makecell[c]{3.145}
        &\makecell[c]{0.410}\\
        
        \makecell[c]{40}
        &\makecell[c]{39}
        &\makecell[c]{0.303}
        &\makecell[c]{2.608}
        &\makecell[c]{3.124}
        &\makecell[c]{0.473}\\
        
        \makecell[c]{30}
        &\makecell[c]{9}
        &\makecell[c]{0.221}
        &\makecell[c]{2.234}
        &\makecell[c]{2.054}
        &\makecell[c]{0.584}\\
        
        \makecell[c]{20}
        &\makecell[c]{19}
        &\makecell[c]{0.248}
        &\makecell[c]{2.380}
        &\makecell[c]{2.064}
        &\makecell[c]{0.784}\\

        \makecell[c]{12}
        &\makecell[c]{3}
        &\makecell[c]{0.202}
        &\makecell[c]{1.976}
        &\makecell[c]{1.984}
        &\makecell[c]{1.302}\\
        
        \makecell[c]{10}
        &\makecell[c]{9}
        &\makecell[c]{0.210}
        &\makecell[c]{2.001}
        &\makecell[c]{2.087}
        &\makecell[c]{1.392}\\
        
        \makecell[c]{1}
        &\makecell[c]{1}
        &\makecell[c]{\textbf{0.131}}
        &\makecell[c]{\textbf{1.220}}
        &\makecell[c]{\textbf{1.934}}
        &\makecell[c]{14.773}\\
        
        \bottomrule
	\end{tabular}
 }
	\caption{Ablation study of different sampling interval at $4 \times$ on PUGAN \cite{li2019pu}. “Distance” means the distance of time steps between $x_1$ and $x_0$. When the sampling interval $=12$, the performance and sampling speed of our methods keep a favorable balance. Meanwhile, we showcase the visual results for the sampling interval $=12$ and $=30$ in Fig \ref{fig_KITTI2}.}
	\label{t121}
\end{table}

\section{Additional Comparative Experiments}

\textbf{Sampling speed.} We conduct the evaluation of sampling speed. Admittedly, as shown in Tab \ref{t211}, our method lags behind in terms of time compared to existing methods. However, a positive aspect is: due to the remarkably strong performance, our method still can maintain state-of-the-art performance even with a certain loss of precision, while also exhibiting a comparable sampling speed (when we set sampling interval $=30$).

\begin{table}[h]
        \scriptsize
  \resizebox{0.48\textwidth}{!}{
	\begin{tabular}{p{1.2cm}p{1.5cm}p{1.5cm}p{1.5cm}p{1.5cm}p{1.5cm}}	
       \bottomrule
  
        \makecell[c]{ }
        &\makecell[c]{PU-Net \cite{yu2018pu}}
        &\makecell[c]{MPU \cite{yifan2019patch}}
        &\makecell[c]{PU-GAN \cite{qian2021pu}}
        &\makecell[c]{Dis-PU \cite{li2021point}}
        &\makecell[c]{PU-EVA \cite{luo2021pu}}\\

        \makecell[c]{Average}
        &\makecell[c]{0.446}
        &\makecell[c]{0.487}
        &\makecell[c]{0.618}
        &\makecell[c]{0.724}
        &\makecell[c]{0.587}\\
        \cline{2-6}

        \makecell[c]{Times(s)$\downarrow$}
        &\makecell[c]{PU-GCN \cite{qian2021pu}}
        &\makecell[c]{NePS \cite{feng2022neural}}
        &\makecell[c]{Grad-PU \cite{he2023grad}}
        &\makecell[c]{Ours}
        &\makecell[c]{I=30}\\

        \makecell[c]{ }
        &\makecell[c]{0.531}
        &\makecell[c]{0.479}
        &\makecell[c]{\textbf{0.403}}
        &\makecell[c]{14.773}
        &\makecell[c]{0.584}\\

        \bottomrule
       
	\end{tabular}
}
	\caption{The results of sampling speed at $4 \times$ on PUGAN \cite{li2019pu}. 'I=30' means that the sampling interval is set to 30 during inference for our method. Our method still maintains the state-of-the-art performance at the comparable sampling speed (For the specific performance, please refer to the table 1 in the main paper).}
	\label{t211}
\end{table}

\begin{figure*}[htp]
	\centering
	\includegraphics[width=\textwidth]{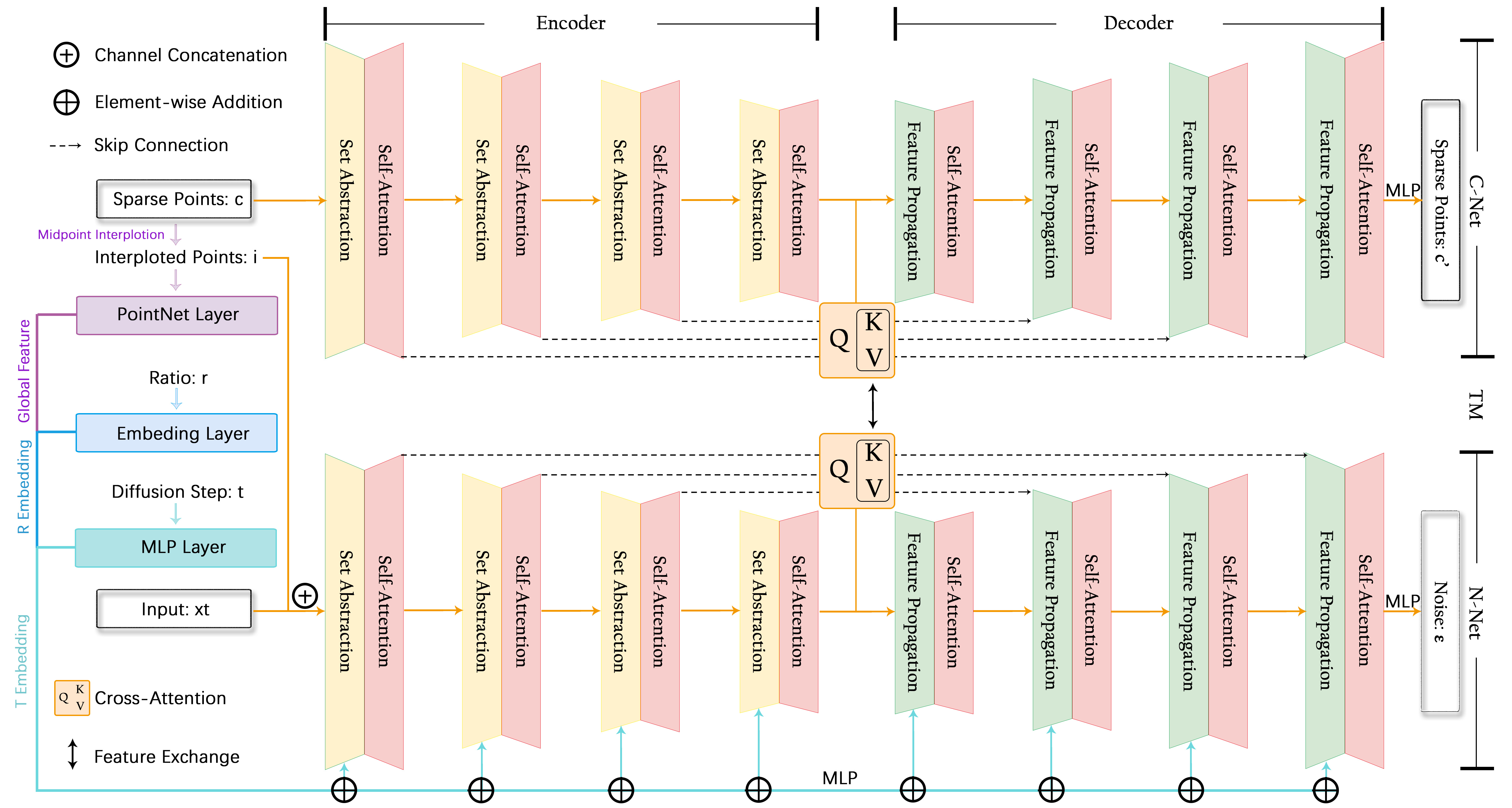}
	\caption{The network detail of PUDM. The encoder of the N-Net and the C-Net consists of set abstraction (SA) layers and self-attention layers. Simultaneously, the decoder consists of feature propagation (FP) layers and self-attention layers. The transfer module used for interaction between the N-Net and the C-Net consists of two cross-attention modules. Compared to the C-Net, the N-Net requires additional information for modeling the diffusion step: the global features, the $R$ embedding, and the $T$ embedding.}
	\label{fig_PUDM}
\end{figure*}

\textbf{Point Cloud Part Segmentation.} We evaluate the quality of point cloud upsampling on point cloud part segmentation. Tab \ref{t221} displays extremely poor results for all point cloud upsampling methods (class accuracy $< 40\%$). 

Existing point cloud upsampling methods (including our methods) are not proficient in semantic-related downstream tasks such as point cloud segmentation and point cloud detection. This is because the semantic order of points is disrupted during the point cloud upsampling process, making each point struggle to align with the semantic label. This greatly limits the application scope of point cloud upsampling in downstream tasks. 

In fact, because the semantic label is uniquely mapped to each point in point clouds, their distributions are inherently similar. Therefore, the model may be able to maintain the semantic order of points during upsampling point clouds.

\begin{table}[h]
        \scriptsize
        \resizebox{0.48\textwidth}{!}{
	\begin{tabular}{c|cccc|cccc}	
        \bottomrule

        \makecell[l]{Datasets}
        &\multicolumn{4}{c}{PointNet \cite{qi2017pointnet} ($\%$)} \vline
        
        &\multicolumn{4}{c}{PointNet++ \cite{qi2017pointnet++} ($\%$)} \\
        \cline{2-5} \cline{6-9}

        \makecell[l]{Methods}
        &\makecell[c]{IA$\uparrow$}
        &\makecell[c]{CA$\uparrow$}
        &\makecell[c]{Im$\uparrow$}
        &\makecell[c]{Cm$\uparrow$}
        
        &\makecell[c]{IA$\uparrow$}
        &\makecell[c]{CA$\uparrow$}
        &\makecell[c]{Im$\uparrow$}
        &\makecell[c]{Cm$\uparrow$}\\
       \hline

        \makecell[l]{Low-res}
        &\makecell[c]{92.16}
        &\makecell[c]{81.12}
        &\makecell[c]{77.13}
        &\makecell[c]{81.11}
        
        &\makecell[c]{92.97}
        &\makecell[c]{84.99}
        &\makecell[c]{81.10}
        &\makecell[c]{83.13}\\

        \makecell[l]{High-res}
        &\makecell[c]{93.47}
        &\makecell[c]{83.05}
        &\makecell[c]{79.01}
        &\makecell[c]{83.99}
        
        &\makecell[c]{94.34}
        &\makecell[c]{86.27}
        &\makecell[c]{82.91}
        &\makecell[c]{85.61}\\
        
        \makecell[l]{PU-Net \cite{yu2018pu}}
        &\makecell[c]{51.92}
        &\makecell[c]{36.05}
        &\makecell[c]{32.69}
        &\makecell[c]{35.79}
        
        &\makecell[c]{52.04}
        &\makecell[c]{36.25}
        &\makecell[c]{32.66}
        &\makecell[c]{35.90}\\

        \makecell[l]{MPU \cite{yifan2019patch}}
        &\makecell[c]{52.01}
        &\makecell[c]{\textbf{36.16}}
        &\makecell[c]{32.76}
        &\makecell[c]{35.89}
        
        &\makecell[c]{52.14}
        &\makecell[c]{36.28}
        &\makecell[c]{32.71}
        &\makecell[c]{36.04}\\       
        
        \makecell[l]{PU-GAN \cite{qian2021pu}}
        &\makecell[c]{52.01}
        &\makecell[c]{35.94}
        &\makecell[c]{\textbf{32.78}}
        &\makecell[c]{35.90}

        &\makecell[c]{52.25}
        &\makecell[c]{36.08}
        &\makecell[c]{32.73}
        &\makecell[c]{36.10}\\

        \makecell[l]{Dis-PU \cite{li2021point}}
        &\makecell[c]{\textbf{52.38}}
        &\makecell[c]{36.02}
        &\makecell[c]{32.79}
        &\makecell[c]{35.96}

        &\makecell[c]{\textbf{52.71}}
        &\makecell[c]{36.16}
        &\makecell[c]{32.78}
        &\makecell[c]{36.16}\\

        \makecell[l]{PU-EVA \cite{luo2021pu}}
        &\makecell[c]{51.80}
        &\makecell[c]{35.94}
        &\makecell[c]{32.69}
        &\makecell[c]{35.82}
        
        &\makecell[c]{52.04}
        &\makecell[c]{36.13}
        &\makecell[c]{32.65}
        &\makecell[c]{35.94}\\

        \makecell[l]{PU-GCN \cite{qian2021pu}}
        &\makecell[c]{51.67}
        &\makecell[c]{35.87}
        &\makecell[c]{32.58}
        &\makecell[c]{35.58}

        &\makecell[c]{51.89}
        &\makecell[c]{36.03}
        &\makecell[c]{32.65}
        &\makecell[c]{35.79}\\

        \makecell[l]{NePS \cite{feng2022neural}}
        &\makecell[c]{51.71}
        &\makecell[c]{35.91}
        &\makecell[c]{32.63}
        &\makecell[c]{35.61}

        &\makecell[c]{52.01}
        &\makecell[c]{36.11}
        &\makecell[c]{32.67}
        &\makecell[c]{35.87}\\

        \makecell[l]{Grad-PU \cite{he2023grad}}
        &\makecell[c]{52.22}
        &\makecell[c]{36.02}
        &\makecell[c]{32.75}
        &\makecell[c]{\textbf{36.07}}

        &\makecell[c]{52.46}
        &\makecell[c]{35.29}
        &\makecell[c]{32.97}
        &\makecell[c]{36.31}\\        
        \hline

        \makecell[l]{Ours}
        &\makecell[c]{51.88}
        &\makecell[c]{36.08}
        &\makecell[c]{32.67}
        &\makecell[c]{36.00}

        &\makecell[c]{52.11}
        &\makecell[c]{\textbf{36.48}}
        &\makecell[c]{\textbf{32.99}}
        &\makecell[c]{\textbf{36.34}}\\          
        \bottomrule
        
	\end{tabular}
}
	\caption{The results of point cloud part segmentation on ShapeNet \cite{chang2015shapenet}. "Low-res" refers to the point cloud subsampled with 512 points, while "High-res" denotes the original test point cloud with 2048 points. Meanwhile, "IA" stands for instance accuracy, and "CA" denotes class accuracy. "Im" means instance mIoU, and "Cm" denotes class mIoU. All methods yield very poor results, as existing point cloud upsampling methods struggle to maintain the semantic order of points.}
	\label{t221}
\end{table}

\section{Implementation}

The network detail of PUDM is shown in Fig \ref{fig_PUDM}. We employ the same training configuration for PUGAN and PU1K. Specifically, we configure batch size $=28$, and conduct $1000$ epochs using an NVIDIA 3090 GPU, taking approximately $5$ days for PUGAN. For the C-Net and the N-Net, the sampling points/the channel dimensions are (1024, 256, 64, 16)/(64, 128, 256, 512) and (1024, 256, 64, 16)/(128, 256, 256, 512), respectively. Meanwhile, in the TM, we set the number $=4$ of head to improve the modeling capacity of our model, and the latent dimension $=64$. In addition, the global features with a dimension of $1024$ are extracted from the interpolated point cloud $i$ through a two-stage PointNet \cite{qi2017pointnet}. The parameters of the $R$ embedding layer are $(256, 128)$, indicating that our method can upsample a point cloud to a maximum of 256 times. The time step $t$ is embedded dimension $=512$ via MLPs \cite{ho2020denoising}. 

\textbf{Encoder.} In the Encoder, the SA first uses iterative farthest point sampling (FPS) to subsample the input points $p_e^l\in \mathbb{R}^{N^l \times 3}$ and the feature $f_e^l \in \mathbb{R}^{N^l \times C_e^l}$ into $p_e^l\in \mathbb{R}^{N^{l+1} \times 3}$ and $f_e^l \in \mathbb{R}^{N^{l+1} \times C_e^l}$ at level $l+1$ ($N^l > N^{l+1}$). Subsequently, we locate $K$ nearest neighbors in $p_e^l$, and aggregate the feature $f_e^l$ and neighbors into $g_{in}\in \mathbb{R}^{N^{l+1} \times C_e^l \times K}$. Next, we further extract the neighborhood features by transforming $g_{in}$ into $g_{out}\in \mathbb{R}^{N^{l+1} \times C_e^{l+1} \times K}$ through MLPs. Simultaneously, to preserve more details, we use the residual connection to aggregate $MLP(g_{in})$ and $g_{out}$. Finally, unlike PointNet++ \cite{qi2017pointnet++} using the max-pooling layer to filter features, we consider using the self-attention layer to retain more fine-grained information \cite{zhao2021point,pan2021variational}, $f_e^{l+1} \in \mathbb{R}^{N^{l+1} \times C_e^{l+1}}$.

\begin{figure*}[htp]
	\centering
	\includegraphics[width=0.9\textwidth]{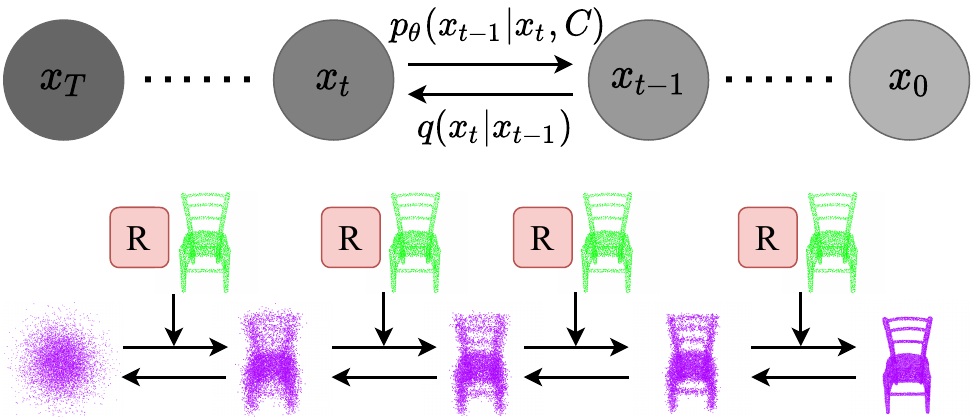}
	\caption{Visualization of the forward process and the reverse process of PUDM. For the forward process, the dense point cloud $x_0$ is gradually added noise according to $q(x_t|x_{t-1})$, until $x_0$ degrades to $x_T$. Simultaneously, for the reverse process, $x_T$ is slowly removed noise according to $p_\theta(x_{t-1}|x_t,C)$, until $x_T$ recovers to $x_0$. We consider adding the conditions at each time step in the reverse process, the sparse point cloud $c$ and the rate prompt $R$, to control the generation of the dense point cloud.}
	\label{fig_DDPM}
\end{figure*}

\textbf{Decoder.} In the Decoder, the FP is similar to the SA, while the FP transforms the input feature $f_d^l \in \mathbb{R}^{N^l \times C_d^l}$ into $f_d^l \in \mathbb{R}^{N^{l+1} \times C_d^l \times K}$ through upsampling ($N_l < N_{l+1}$). Subsequently, we feed $f_d^l$ into a self-attention layer to obtain $f_d^l \in \mathbb{R}^{N^{l+1} \times C_d^l}$. Simultaneously, to propagate features, we aggregate $f_d^l$ with the points $p_d^l$ and the features $f_e^l$ from the SA at the same level. Finally, we transform $f_d^l$ into $f_d^{l+1} \in \mathbb{R}^{N^{l+1} \times C_d^{l+1}}$ through MLPs.

\section{Formula Derivation of DDPM for PCU}
In this section, we provide the theoretical foundation for the application of DDPM in PCU. Due to the page limitations, our derivation process focuses more on the overall logic, overlooking some details.

\textbf{The forward process.}  The forward process $q$ is modeled as a Markov chain, while the each step follows an independent Gaussian distribution. This gradually adds noise to $x$ until $x$ degrades to $z$. The process is irrelevant of the condition $c$ (i.e. the sparse point cloud). Formally, given a time step $t \sim \mathcal{U}(T)$ and the dense point cloud $x_0 \sim P_{data}$, we can compute the forward process by the conditional distribution $q(x_{1:T}|x_0)$:


\begin{equation}
\begin{split}
    \label{f41}
    q(x_{1:T}|x_0)= \frac{q(x_{0:T})}{q(x_0)} \quad\quad\quad\quad\quad\quad\quad\quad\quad\quad\quad\quad\quad\quad\;\;\\
= \frac{\textcolor{red}{q(x_T|x_{0:T-1})}q(x_{0:T-1})}{q(x_0)}\quad\quad\quad\quad\quad\quad\quad\quad\\
\textcolor{blue}{Markov \; Property:} \quad\quad\quad\quad\quad\quad\quad\quad\quad\quad\quad\quad\quad\quad\quad\\ 
= \frac{\textcolor{red}{q(x_T|x_{T-1})}q(x_{T-1}|x_{0:T-2})q(x_{0:T-2})}{q(x_0)}\quad\quad\\ 
= \frac{q(x_T|x_{T-1})q(x_{T-1}|x_{T-2})...q(x_1|x_0)\textcolor{red}{\cancel{q(x_0)}}}{{\textcolor{red}{\cancel{q(x_0)}}}}\\
= \prod_{t=1}^{T}q(x_t|x_{t-1})\quad\quad\quad\quad\quad\quad\quad\quad\quad\quad\quad\quad
\end{split}
\end{equation}
where $q(x_t|x_{t-1}) = \mathcal{N}(x_t; \sqrt{1-\beta_t}x_{t-1},\beta_tI)$.  $\beta_t$ is a predefined and increasing variance term ($\beta_t \in [0.0001,0.02]$ in this paper). 


Meanwhile, to enable the sampling-differentiable training, we utilize the reparameterization  trick\cite{ho2020denoising} : $x_t=\mu+\sigma\epsilon_t$,  $\epsilon_t \sim \mathcal{N}(\epsilon_t; 0, I)$. Next, we can obtain a more simplified formulation of computing $x_t$ by setting $\alpha_t = 1 - \beta_t$, and $\overline{\alpha}_t = \prod_{t=1}^{T}\alpha_t$:

\begin{equation}
\begin{split}
	\label{f42}
x_t=\sqrt{1-\beta_t}x_{t-1} + \beta_t\epsilon_t\quad\quad\quad\quad\quad\quad\quad\quad\quad\quad\quad\quad\;\\
=\sqrt{\alpha_t}x_{t-1} + \sqrt{1-\alpha_t}\epsilon_t\quad\quad\quad\quad\quad\quad\quad\quad\quad\quad\;\;\;\;\;\\
=\sqrt{\alpha_t}(\sqrt{\alpha_{t-1}}x_{t-2} + \sqrt{1-\alpha_{t-1}}\epsilon_{t-1}) + \sqrt{1-\alpha_t}\epsilon_t\\ 
=\sqrt{\alpha_t\alpha_{t-1}}x_{t-2}+\textcolor{red}{\sqrt{\alpha_t-\alpha_t\alpha_{t-1}}\epsilon_{t-1}+\sqrt{1-\alpha_t}\epsilon_t} \\
=\sqrt{\alpha_t\alpha_{t-1}}x_{t-2}+\textcolor{red}{\sqrt{1-\alpha_t\alpha_{t-1}}\epsilon}\quad\quad\quad\quad\quad\quad\quad\;\;\\
...\quad\quad\quad\quad\quad\quad\quad\quad\quad\quad\quad\quad\quad\quad\quad\quad\quad\quad\quad\quad\quad\\
=\sqrt{\overline{\alpha}_t}x_0 + \sqrt{1-\overline{\alpha}_t} \epsilon\quad\quad\quad\quad\quad\quad\quad\quad\quad\quad\quad\;\;\quad
\end{split}
\end{equation}
where $\epsilon$ represents the combination of multiple Gaussian noise terms.

Therefore, $x_t$ is only related to the dense point cloud $x_0$ and the time step $t$ in the forward process.

\textbf{The reverse process.} Similarly, the reverse process $p$ is also modeled as a Markov chain, while the each step is assumed to follow an independent Gaussian distribution. This slowly removes noise from $z$ until $z$ recovers to $x$. Formally, given a set of conditions $C=\{c_i|i=1..S\}$ ("$S$" means the number of conditions), we can compute the reverse process by the joint distribution $p_\theta(x_{0:T},C)$:

\begin{equation}
\begin{split}
	\label{f43}
p_\theta(x_{0:T},C)=\textcolor{red}{p_\theta(x_0|x_{1:T},C)}p_\theta(x_{1:T},C)\quad\quad\quad\quad\quad\quad\;\;\\
\textcolor{blue}{Markov \; Property:} \quad\quad\quad\quad\quad\quad\quad\quad\quad\quad\quad\quad\quad\quad\quad\\
=\textcolor{red}{p_\theta(x_0|x_1,C)}p_\theta(x_1|x_{2:T},C)p_\theta(x_{2:T},C)\quad\quad\quad\\ 
\quad\quad=p_\theta(x_0|x_1,C)...p_\theta(x_T|x_{T-1},C)p(x_T,C)\quad\;\;\;\;\;\;\\ 
=p(x_T,C)\prod_{t=1}^{T}p_\theta(x_{t-1}|x_t,C)\quad\quad\quad\quad\quad\quad\quad\;\;\\ 
\textcolor{blue}{x_T \sim \mathcal{N}(x_T;0,I)}\quad\quad\quad\quad\quad\quad\quad\quad\quad\quad\quad\quad\quad\quad\quad\quad\\ 
=p(x_T)\prod_{t=1}^{T}p_\theta(x_{t-1}|x_t,C)\quad\quad\quad\quad\quad\quad\quad\quad\;\;\\
\end{split}
\end{equation}

When $C=\emptyset$, Eq \ref{f43} transforms into the reverse process of standard DDPM (i.e. unconditional DDPM).

Then, as DDPM is modeled to be reversible, we can directly compute the posterior distribution $q(x_{t-1}|x_t,x_0)$ in the forward process:

%

\begin{equation}
\begin{split}
\label{f44}
q(x_t|x_{t-1},x_0)=\frac{\textcolor{red}{q(x_{t-1}|x_t,x_0)}q(x_t,x_0)}{q(x_{t-1},x_0)}\quad\quad\quad\\
\textcolor{red}{q(x_{t-1}|x_t,x_0)}=\frac{q(x_t|x_{t-1},x_0)q(x_{t-1}|x_0)}{q(x_t|x_0)}\quad\quad\\
\textcolor{blue}{Markov \; Property:} \quad\quad\quad\quad\quad\quad\quad\quad\quad\quad\quad\quad\quad\quad\quad\\
=\frac{\textcolor{red}{q(x_t|x_{t-1})}q(x_{t-1}|x_0)}{q(x_t|x_0)}\quad\quad\;\;\;\;\;
\end{split}
\end{equation}
where $q(x_t|x_{t-1}) = \mathcal{N}(x_t;\sqrt{\alpha_t}x_{t-1},(1-\alpha_t)I)$, $q(x_{t-1}|x_0) = \mathcal{N}(x_{t-1};\sqrt{\overline{\alpha}_{t-1}}x_0,(1-\overline{\alpha}_{t-1})I)$, and $q(x_t|x_0) = \mathcal{N}(x_t;\sqrt{\overline{\alpha}_t}x_0,(1-\overline{\alpha}_t)I)$. In fact, each time step in the reverse process aims to gradually fit the posterior distribution under corresponding time step in the forward process (the posterior distribution represents the inverse process of the forward process, not the reverse process of DDPM), i.e. $p_\theta(x_{t-1}|x_t,C) \approx q(x_{t-1}|x_t,x_0)$. Therefore, deriving $p_\theta(x_{t-1}|x_t,C)$ is equated to deriving $q(x_{t-1}|x_t,x_0)$.

Subsequently, by substituting $q(x_t|x_{t-1})$, $q(x_{t-1}|x_0)$ and $q(x_t|x_0)$ into Eq \ref{f44}, we obtain the mean $\widetilde{\mu_t}$ and the variance $\widetilde{\sigma_t}$ of $q(x_t|x_{t-1})$:


\begin{equation}
\begin{split}
\label{f45}
\widetilde{\mu_t}=\frac{\sqrt{\alpha_t}(1-\overline{\alpha}_{t-1})}{1-\overline{\alpha_t}}x_t+\frac{\sqrt{\overline{\alpha}_{t-1}}(1-\alpha_t)}{1-\overline{\alpha}_t}x_0\\
\widetilde{\sigma}_t=\frac{1-\overline{\alpha}_{t-1}}{1-\overline{\alpha}_t}(1-\alpha_t)\quad\quad\quad\quad\quad\quad\quad\quad\;\;\;
\end{split}
\end{equation}

We can clearly realize that $\widetilde{\sigma}_t$ is a constant. 

Next, we substitute $x_0=\frac{x_t-\sqrt{1-\overline{\alpha}_t}\epsilon}{\sqrt{\overline{\alpha}_t}}$ into $\widetilde{\mu_t}$ to the new expression of the mean $\widetilde{\mu_t}$:

\begin{equation}
\begin{split}
\label{f46}
\widetilde{\mu_t}=\frac{1}{\sqrt{\alpha_t}}(x_t-\frac{1-\alpha_t}{\sqrt{1-\overline{\alpha}_t}}\epsilon)
\end{split}
\end{equation}

Therefore, for the posterior distribution $q(x_{t-1}|x_t,x_0)$, we can compute $x_{t-1}$ solely by providing $x_t$ and $\epsilon$ (where $x_0$ must be considered as prior knowledge in forward process). 

Although according to Eq \ref{f45} and Eq \ref{f46}, the posterior distribution $q(x_{t-1}|x_t,x_0)$ is known, we can not directly utilize it to deriving $x_0$ due to involving $q(x_t|x_0)$, which requires obtaining $x_t$ and $\epsilon$ from the forward process.

Typically, the network $f(x_t,t)$ fits $\epsilon$ during the training process of DDPM, as $x_t$ is known in the reverse process ($x_t \sim \mathcal{N}(x_t;0,I)$). Simultaneously, to introduce the condition set $C$ during in the reverse process (Fig \ref{fig_DDPM}), the network increases additionally inputs, i.e. $f(x_t,t,C)$. In PUDM, the condition set $C = \{c,r\}$ represent the sparse point cloud and the rate prompt between the sparse point cloud and the dense point cloud.

\textbf{Training objective under specific conditions.} The training objective of DDPM under specific conditions is to maximize a \textbf{E}vidence \textbf{L}ower \textbf{BO}und (ELBO), due to directly optimize the log-likelihood $\log p_\theta(x_0,C)$ is intractable. 

We directly convert the log-likelihood $\log p_\theta(x_0,C)$ into a loss form $-\log p_\theta(x_0,C)$. We first add a KL divergence item $D_{kl}[q(x_{1:T}|x_0)||p_\theta(x_{1:T}|x_0,C)]$ to $-\log p_\theta(x_0,C)$:

\begin{equation}
\begin{split}
	\label{f47}
-\log p_\theta(x_0,C) \quad\quad\quad\quad\quad\quad\quad\quad\quad\quad\quad\quad\quad\quad\quad\quad\;\;\;\\
\leq-\log p_\theta(x_0,C)+D_{kl}[q(x_{1:T}|x_0)||(p_\theta(x_{1:T}|x_0,C))]\;\;\\
\leq -\log p_\theta(x_0,C)+\quad\quad\quad\quad\quad\quad\quad\quad\quad\quad\quad\quad\quad\quad\quad\\ 
\int q(x_{1:T}|x_0)\log \frac{q(x_{1:T}|x_0)}{p_\theta(x_{1:T}|x_0,C)}dx_{1:T}\quad\quad\quad\quad\quad\quad\\ 
\leq -\log p_\theta(x_0,C)+\quad\quad\quad\quad\quad\quad\quad\quad\quad\quad\quad\quad\quad\quad\quad\\ 
\int q(x_{1:T}|x_0)\log \frac{q(x_{1:T}|x_0)}{\frac{p_\theta(x_{1:T},x_0,C)}{p_\theta(x0,C)}}dx_{1:T}\quad\quad\quad\quad\quad\quad\;\;\\ 
\leq -\log p_\theta(x_0,C)+\quad\quad\quad\quad\quad\quad\quad\quad\quad\quad\quad\quad\quad\quad\quad\\
\int q(x_{1:T}|x_0)(\log \frac{q(x_{1:T}|x_0)}{p_\theta(x_{0:T},C)}+\textcolor{red}{\log p_\theta(x_0,C)}) dx_{1:T}\\
\leq -\log p_\theta(x_0,C)+\quad\quad\quad\quad\quad\quad\quad\quad\quad\quad\quad\quad\quad\quad\quad\\
\mathbb{E}_{q({x_{1:T|x_0}})}\log \frac{q(x_{1:T}|x_0)}{p_\theta(x_{0:T},C)}+\textcolor{red}{\log p_\theta(x_0,C)}\quad\quad\quad\;\\
\leq\mathbb{E}_{q({x_{1:T|x_0}})}\log \frac{q(x_{1:T}|x_0)}{p_\theta(x_{0:T},C)}\quad\quad\quad\quad\quad\quad\quad\quad\quad\quad\quad\;\\
\textcolor{blue}{Adding\;\mathbb{E}_{q({x_0})}\;to\;the\;both\;sides:} \quad\quad\quad\quad\quad\quad\quad\quad\quad\quad\quad\\
	\textcolor{red}{-\mathbb{E}_{q({x_0})}}\log p_\theta(x_0,C) \leq 
\textcolor{red}{\mathbb{E}_{q(x_{0:T})}}\log\frac{q(x_{1:T}|x_0)}{p_\theta(x_{1:T},C)} \quad\quad\quad\;
\end{split}
\end{equation}

Then, leveraging the Markov property and Bayes' theorem, we can obtain the loss form $L_{ELBO}$ of the ELBO:

\begin{equation}
\begin{split}
\label{f48}
 L_{ELBO} = \mathbb{E}_{q}\frac{q(x_{1:T}|x_0)}{p_\theta(x_{1:T},C)}\; \;  \quad \quad \quad \quad \quad\quad\quad\quad\quad\quad\quad\;\;\\
= \underbrace{\mathbb{E}_{q}[D_{KL}(q(x_T|x_0)||p(x_T))}_{\substack{\color{orange}{\Circled{1}}}}
\quad\quad\quad\quad\quad\quad\quad\\
+\underbrace{\sum_{t=2 \in T} D_{KL}(q(x_{t-1}|x_t,x_0)||p_\theta(x_{t-1}|x_t,C))}_{\substack{\color{red}{\Circled{2}}}}\\
- \underbrace{\log p_\theta (x_0|x_1,C)}_{\substack{\color{blue}{\Circled{3}}}}] \quad \quad \quad \quad \quad\quad\quad\quad \quad\quad\quad
\end{split}
\end{equation}

Subsequently, we can eliminate the constant term $\color{orange}{\Circled{1}}$, and combine $\color{red}{\Circled{2}}$ and $\color{blue}{\Circled{3}}$  (where $\color{red}{\Circled{2}} = \color{blue}{\Circled{3}}$, when $t=1$) to obtain a more simplified expression:

\begin{equation}
\begin{split}
\label{f49}
L_{ELBO}=\quad\quad\quad\quad\quad\quad\quad\quad\quad\quad\quad\quad\quad\quad\quad\quad\quad\quad\quad\quad\quad\quad\;\;\;\;\\
\sum_{t=1 \in T} D_{KL}(q(x_{t-1}|x_t,x_0)||p_\theta(x_{t-1}|x_t,C))\quad\quad\quad\quad\quad\quad
\end{split}
\end{equation}

In order to approximate $p_\theta(x_{t-1}|x_t,C)$ to $q(x_{t-1}|x_t,x_0)$, we represent the mean $\mu_\theta(x_t,t,C)$ through a neural network with parameter $\theta$. Meanwhile, we can further expand $L_{ELBO}$ to obtain a more simplified form:

\begin{equation}
\begin{split}
    \label{f410}
q(x_{t-1}|x_t,x_0) \sim 
\mathcal{N}(x_{t-1};\frac{1}{\sqrt{\alpha_t}}(x_t-\frac{1-\alpha_t}{\sqrt{1-\overline{\alpha}_t}}\epsilon), \widetilde{\sigma}_tI)\quad\quad\quad\;\;\;\;\;\\    
p_\theta(x_{t-1}|x_t,C) \sim \mathcal{N}(x_{t-1};\mu_\theta(x_t,t,C), \widetilde{\sigma}_tI)\quad\quad\quad\quad\quad\quad\quad\;\;\;\;\;\;\\    
\textcolor{blue}{where \; \widetilde{\sigma}_t = \frac{1-\overline{\alpha}_{t-1}}{1-\overline{\alpha}_t}(1-\alpha_t)}\quad\quad\quad\quad\quad\quad\quad\quad\quad\quad\quad\quad\quad\;\;\;\;\;\;\;\;\\
L_{ELBO}\quad\quad\quad\quad\quad\quad\quad\quad\quad\quad\quad\quad\quad\quad\quad\quad\quad\quad\quad\quad\quad\quad\quad\;\;\;\;\;\;\\
=\mathbb{E}_{q(x_{0:T})}(\frac{1}{2\widetilde{\sigma}_t^2}||\widetilde{\mu_t}-\mu_\theta(x_t,t,C)||^2)\quad\quad\quad\quad\quad\quad\quad\quad\quad\quad\quad\;\;\\
=\mathbb{E}_{q(x_{0:T}),\epsilon}(\frac{1}{2\widetilde{\sigma}_t^2}||(\frac{1}{\sqrt{\alpha_t}}(x_t-\frac{1-\alpha_t}{\sqrt{1-\overline{\alpha}_t}}\epsilon)-\quad\quad\quad\quad\quad\quad\quad\quad\quad\\ \mu_\theta(x_t,t,C)||^2)\quad\quad\quad\quad\quad\quad\quad\quad\quad\quad\quad\quad\quad\quad\quad\quad\quad\quad\quad\quad\;\;\\
=\mathbb{E}_{q(x_{0:T}),\epsilon}(\frac{(1-\alpha_t)^2}{2\widetilde{\sigma}_t^2\alpha_t(1-\overline{\alpha}_t)}||(\frac{1}{\sqrt{\alpha_t}}(x_t-\frac{1-\alpha_t}{\sqrt{1-\overline{\alpha}_t}}\epsilon))-\quad\quad\quad\quad\;\\
(\frac{1}{\sqrt{\alpha_t}}(x_t-\frac{1-\alpha_t}{\sqrt{1-\overline{\alpha}_t}}\epsilon_\theta(x_t,t,C)))||^2)\quad\quad\quad\quad\quad\quad\quad\quad\quad\quad\quad\quad\\
=\mathbb{E}_{q(x_{0:T}),\epsilon}||\epsilon-\epsilon_\theta(\textcolor{red}{x_t},t,C)||^2\quad\quad\quad\quad\quad\quad\quad\quad\quad\quad\quad\quad\quad\quad\;
\end{split}
\end{equation}

Next, given $t \sim \mathcal{U}(T)$ and $\epsilon \sim \mathcal{N}(0,I)$, we can obtain the training objective $L(\theta)$ for DDPM under specified conditions:

\begin{equation}
\begin{split}
	\label{f411}
L(\theta) =\quad\quad\quad\quad\quad\quad\quad\quad\quad\quad\quad\quad\quad\quad\quad\quad\quad\quad\quad\quad\\
 \mathbb{E}_{t \sim U(T), \epsilon \sim \mathcal{N}(0,I)}||\epsilon - \epsilon_\theta(\textcolor{red}{\sqrt{1-\overline{\alpha}_t} \epsilon + \sqrt{\overline{\alpha}_t}x_0},t,C)||^2 
\end{split}
\end{equation}

Similar to the reverse process, when $C = \emptyset$, $L(\theta)$ means the training objective of the standard DDPM.

We provide a general derivation process concerning both conditional and unconditional DDPM. Therefore, this is not only applicable to PUDM but also to other tasks employing DDPM.

\section{More Visualization}
\begin{figure*}[htp]
	\centering
	\includegraphics[width=0.83\textwidth]{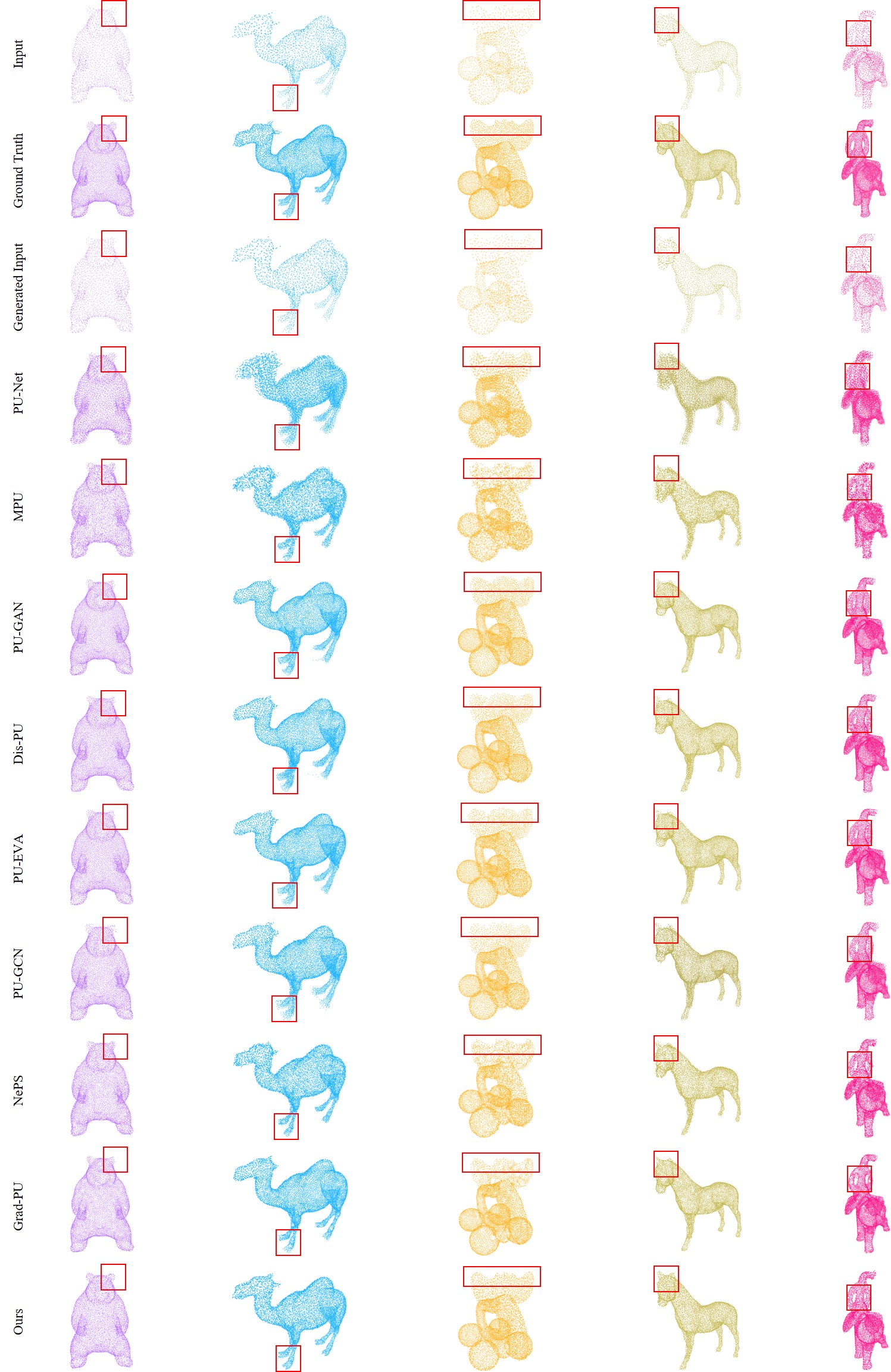}
	\caption{Visualization of point cloud upsampling at $4 \times$ on PUGAN \cite{li2019pu}.}
	\label{fig_pugan}
\end{figure*}

\begin{figure*}[htp]
	\centering
	\includegraphics[width=1.0\textwidth]{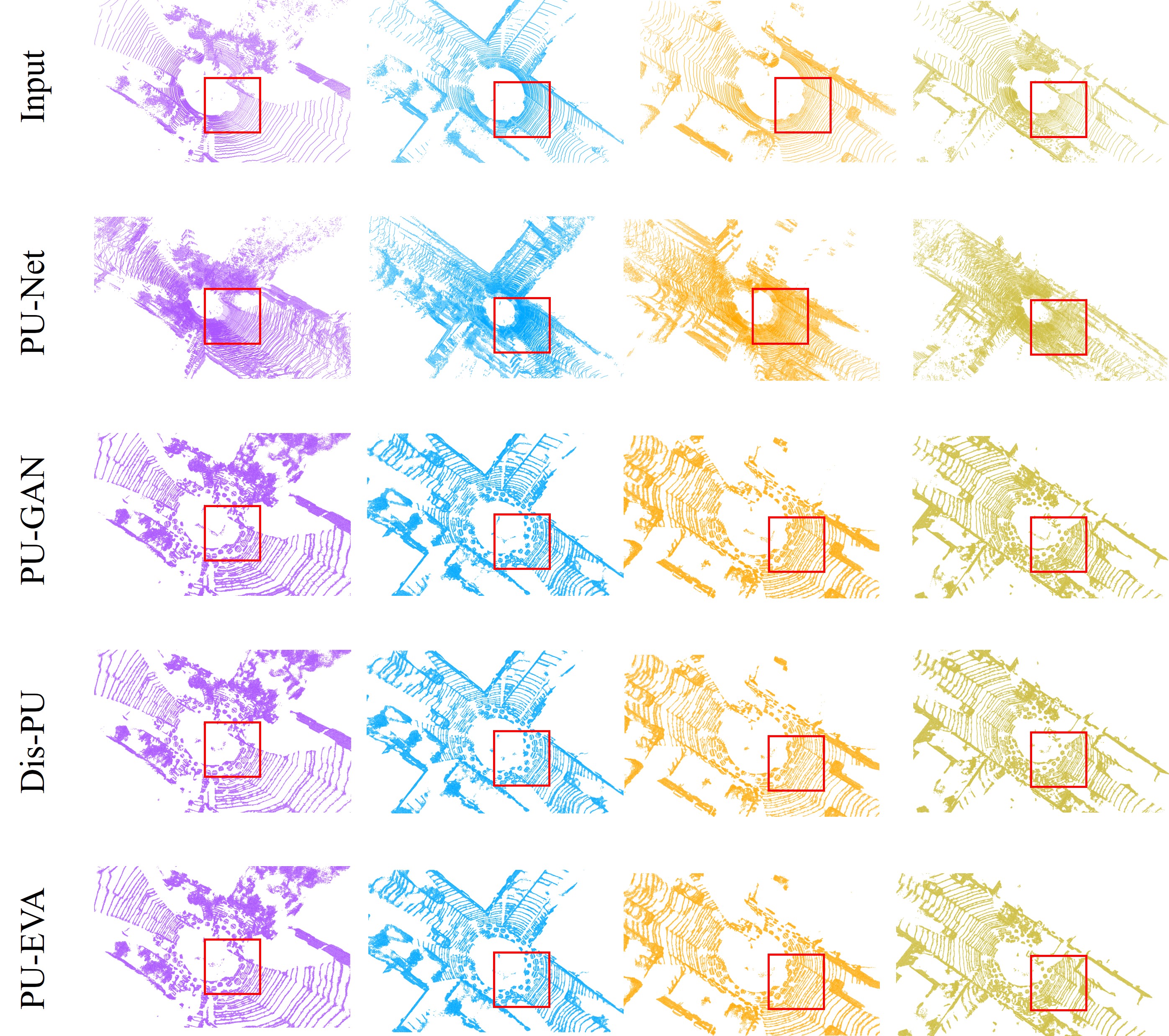}
	\caption{Visualization of point cloud upsampling at $4 \times$ on KITTI \cite{geiger2013vision}.}
	\label{fig_KITTI1}
\end{figure*}

\begin{figure*}[htp]
	\centering
	\includegraphics[width=1.0\textwidth]{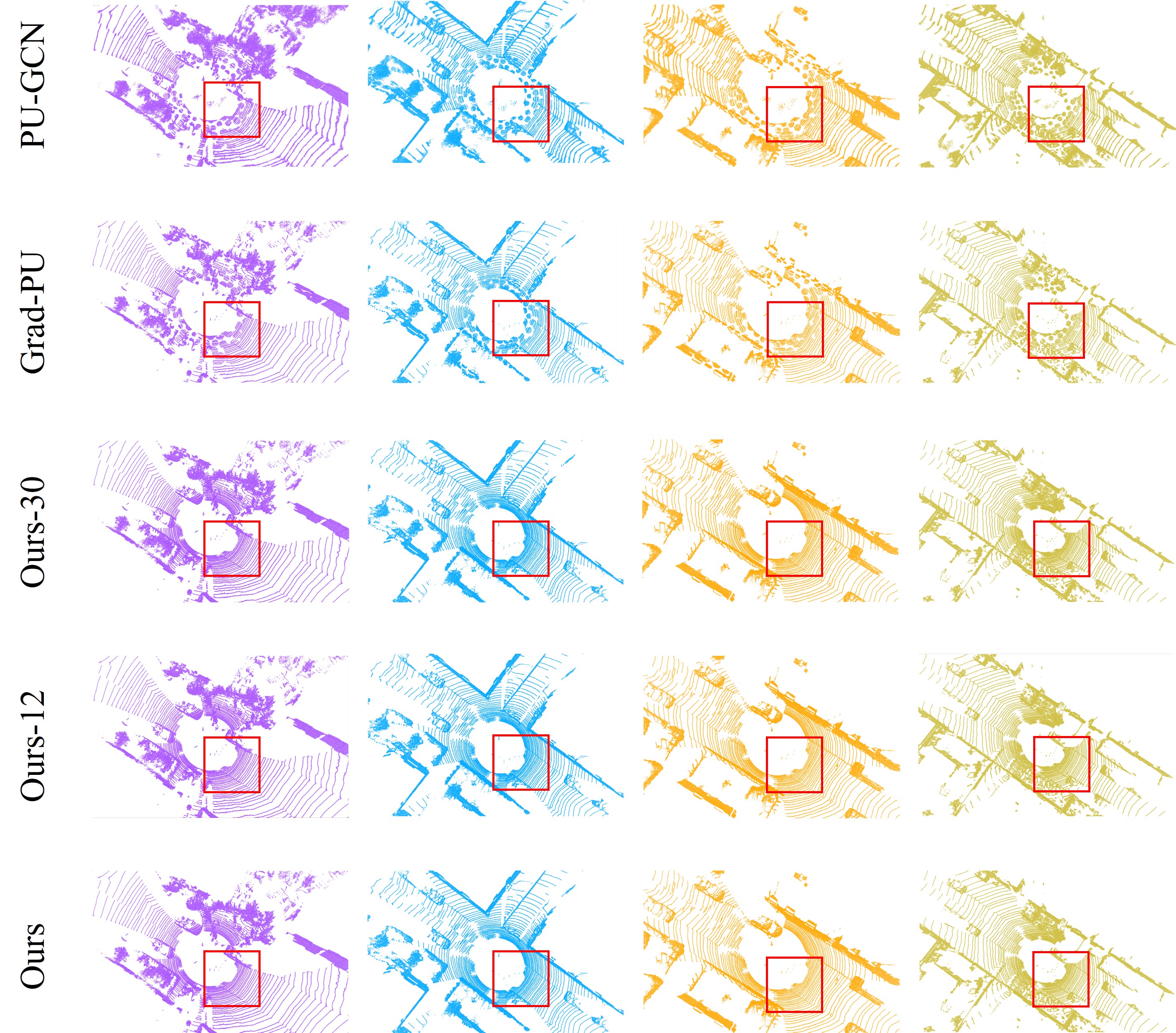}
	\caption{Visualization of point cloud upsampling at $4 \times$ on KITTI \cite{geiger2013vision}. "Ours-12" and "Ours-30" represent the results of our method at sampling intervals $=12$ and $=30$, respectively.}
	\label{fig_KITTI2}
\end{figure*}

We display additional visual results of upsampled point clouds in Fig \ref{fig_pugan}, Fig \ref{fig_KITTI1} and Fig \ref{fig_KITTI2}. 


\end{document}